\documentclass[11pt]{article}
\usepackage[in]{fullpage}

  \usepackage[pdftex]{graphicx}
\usepackage{placeins}
\usepackage{commath}
\usepackage{comment}
\usepackage{float}
\usepackage{booktabs} 
\usepackage{color,soul}
\usepackage[font=small,skip=0pt]{subcaption}
\usepackage[font=small]{caption}
\usepackage{array}
\usepackage{amssymb}
\usepackage{bm,bbm}
\newcommand{\1}{\mathbf{1}}
\newcommand{\0}{\mathbf{0}}
\usepackage{amsopn}
\renewcommand{\a}{\mathbf{a}}
\newcommand{\bb}{\mathbf{b}}
\renewcommand{\c}{\mathbf{c}}
\newcommand{\e}{\mathbf{e}}

\newcommand{\g}{\mathbf{g}}
\newcommand{\h}{\mathbf{h}}

\renewcommand{\r}{\mathbf{r}} 

\renewcommand{\u}{\mathbf{u}}
\renewcommand{\v}{\mathbf{v}}
\newcommand{\w}{\mathbf{w}}

\newcommand{\y}{\mathbf{y}}
\newcommand{\z}{\mathbf{z}}
\usepackage{mathtools}
\usepackage{makecell}
\usepackage{authblk}
\newcommand{\ackname}{Acknowledgements}
\newcommand{\ind}{\mathbbm{1}}

\newcommand\I{\mathbb{I}}

\newcommand\R{\mathbb{R}}

\newtheorem{theorem}{Theorem}

\DeclareMathOperator{\argmin}{argmin}
\DeclareMathOperator{\up}{up}
\DeclareMathOperator{\down}{down}

\newcommand{\remove}[1]{ }
\newcommand{\TempRemove}[1]{ }

\usepackage{amsmath}
\usepackage{array}
\begin{document}
%
\title{High-Dimensional Dependency Structure Learning\\ for Physical Processes}

%

\author[1]{Jamal Golmohammadi}
\author[2]{Imme Ebert-Uphoff}
\author[1]{Sijie He}
\author[3]{Yi Deng}
\author[1]{Arindam Banerjee}
\affil[1]{Department of Computer Science \& Engineering, University of Minnesota, Twin-cities\\

Email: golmo002@umn.edu, hexxx893@umn.edu, banerjee@cs.umn.edu}
\affil[2]{Department of Electrical and Computer Engineering, Colorado State University\\

Email: iebert@colostate.edu}
\affil[3]{School of Earth and Atmospheric Sciences, Georgia Institute of Technology\\

Email: yi.deng@eas.gatech.edu}



\maketitle
\begin{abstract}
In this paper, we consider the use of structure learning methods for probabilistic graphical models to identify statistical dependencies in high-dimensional physical processes. Such processes are often synthetically characterized using PDEs (partial differential equations) and are observed in a variety of natural phenomena, including geoscience data capturing atmospheric and hydrological phenomena. Classical structure learning approaches such as the PC algorithm and variants are challenging to apply due to their high computational and sample requirements. Modern approaches, often based on sparse regression and variants, do come with finite sample guarantees, but are usually highly sensitive to the choice of hyper-parameters, e.g., parameter $\lambda$ for sparsity inducing constraint or regularization. In this paper, we present ACLIME-ADMM, an efficient two-step algorithm for adaptive structure learning, which estimates an edge specific parameter $\lambda_{ij}$ in the first step, and uses these parameters to learn the structure in the second step. Both steps of our algorithm use (inexact) ADMM to solve suitable linear programs, and all iterations can be done in closed form in an efficient block parallel manner. We compare ACLIME-ADMM with baselines on both synthetic data simulated by partial differential equations (PDEs) that model advection-diffusion processes, and real data (50 years) of daily global geopotential heights to study information flow in the atmosphere. ACLIME-ADMM is shown to be efficient, stable, and competitive, usually better than the baselines especially on difficult problems. On real data, ACLIME-ADMM recovers the underlying structure of global atmospheric circulation, including switches in wind directions at the equator and tropics entirely from the data.

\end{abstract}


%

\section{Introduction}

The ability to infer interactions between variables from high-dimensional data sets has the potential to help geoscientists answer numerous questions critical for improved modeling and prediction capabilities for various geoscience processes.  Using atmospheric science as an example, it would enable us to (1) delineate better the interactions between atmospheric disturbances of different spatial scales, which is critical for understanding the working of a weather-climate continuum; (2) develop a better understanding of the degree and spatial pattern of coupling between the top of atmosphere (TOA) radiative imbalance and surface temperatures, which provides a unique perspective of climate feedback processes; (3) identify causal pathways in the atmospheric circulation and infer how they might change under a warming climate \cite{DeEb:GRL2014}; and (4) study the dynamical processes of air-sea interaction that lead to the onset of the monsoons. These applications would contribute to both our understanding of the key processes determining the main features of the Earth's climate system and our capabilities to predict changes in this system with changing external forcing (e.g., aerosols and greenhouse gas emissions) in the near future.

Structure learning is thus emerging in the geosciences as an important tool for that purpose.
%
Recent applications include the study of tele-connections \cite{chu2005data} and the study of atmospheric information flow around the globe \cite{EbDe:GRL2012}.  
Such studies have only recently become possible, thanks to increasing computational power, combined with the rapidly increasing amount of observational and model output data for the earth atmosphere \cite{IPCC:13}.

\subsection{State-of-the-art and Its Limitations} 
Structure learning methods can be broadly divided into two groups. The first group of methods were developed in the seminal work by Pearl \cite{pearl2009causality} and Spirtes-Glymour-Scheines \cite{spirtes2000causation}, among others. The PC algorithm and its variants \cite{spirtes2000causation}, \cite{spirtes1991algorithm}, \cite{kalisch2007estimating}, \cite{harris2013pc} constitute the most popular methods from this family, and are capable of producing the skeletal structure of the underlying Bayesian network capturing the data dependency. However, such methods are `information-theoretic' in the sense that they give the correct output in the asymptotic limit of infinite samples \cite{kalisch2007estimating} and may need exponential computation in the worst case. On the statistical side, in the real world setting of finite samples, such methods cannot (yet) characterize the probability of error (or p-value) of the graph produced. On the computational side, while advances have been made \cite{colombo2014order}, existing advanced implementations of the PC algorithm do not scale beyond ~100,000 variables, whereas geoscience data routinely involves higher dimensional physical processes \cite{ML-GEOchallenges}.

The second group of methods, such as graphical Lasso \cite{friedman2008sparse}, \cite{meinshausen2006high} and CLIME \cite{cai2011constrained}, have seen active development over the past decade \cite{cai2016estimating}, \cite{wang2013large} and come with rigorous finite sample statistical guarantees and efficient computational algorithms. However, such algorithms do need to assume the joint distribution over the variables to be of a specific (semi)parametric family, e.g., multivariate Gaussian (copula), Ising, multivariate Poisson, etc. 
The second group of methods \cite{drton2016structure}, based on sparse high-dimensional estimation, can do structure learning 
by estimating the moral graph of the underlying Bayes net using finite samples {\em in theory}, but has a major limitation {\em in practice}: {\em instability} due to (hyper-)parameter choices. Such methods, based on Lasso and variants need to choose constants, say $\lambda$ for Lasso \cite{friedman2008sparse}, \cite{meinshausen2006high}, which determine the level of sparsity. For structure learning, the output graph can vary significantly based on the specific parameters used. Recent years have seen advances on making the output more stable possibly by repeatedly running the algorithm for different values of the parameters possibly on (disjoint) subsets of the sample \cite{meinshausen2010stability}, \cite{meinshausen2006high}. Such advances, while promising, are computationally demanding, due to the need for repeated runs, and can be statistically demanding due to the need for larger samples.

\subsection{Contributions of This Work} 
We seek to address the issues of both stability and computational demands in this work through the following
key contributions. 
First, we introduce  ACLIME-ADMM, an efficient two-step algorithm for adaptive structure learning, which estimates an edge specific parameter $\lambda_{ij}$ for edge $(i,j)$ in the first step, and uses these parameters to learn the structure in the second step. Both steps of our algorithm use (inexact) ADMM to solve suitable linear programs, and all iterations can be done in closed form. Second, we propose a significantly more scalable version of ACLIME-ADMM based on block updates rather than in single column updates for basic ACLIME-ADMM. The block updates are non-trivial since every column solves a mildly different linear program. The proposed method is developed based on a careful analysis of the shared structure of these problems, and first does a block update followed by column specific adjustments. Third, we illustrate the effectiveness of ACLIME-ADMM by comparisons with state-of-the-art baselines, i.e, PC-variants (PC stable \cite{colombo2014order}) and CLIME variants (CLIME-ADMM \cite{wang2013large}) through extensive experiments on both synthetic and real data involving geo-physical processes. Furthermore, methods from {\it structure learning} for probabilistic graphical models \cite{Pearl:book,drton2016structure} have been applied with great success in disciplines ranging from social sciences \cite{spirtes2000causation} to bioinformatics \cite{ chen2006effective},
to identify {\it direct} dependencies. The proposed algorithm can also be applied in such area with its advantages of efficiency and scaliablity.
\remove{
The synthetic data is generated from two-dimensional advection-diffusion processes captured by suitable partial differential equations (PDEs) \cite{Bergman_book}. We consider several scenarios including circular, ring, and cross-current flow, with data generated from the corresponding PDEs being fed into our structure learning algorithms. Through both qualitative and quantitative results, ACLIME-ADMM is shown to be better than the baselines in all settings, and the performance is especially strong in more challenging scenarios. For real data, we use gridded global geopotential height data at 500mb, i.e., at any location, the height at which air pressure measures 500mb. We use global data from 1950-2000 at daily resolution with focus on the boreal winter (Dec, Jan, Feb). Using such global daily pressure variation data over 50 years, ACLIME-ADMM is able to recover the underlying structure of global atmospheric circulation, including switches in wind directions at the equator and tropics entirely from the data. The result is remarkable because of the stability and robustness illustrated by ACLIME-ADMM to day-to-day local pressure fluctuations and other variations. Overall, ACLIME-ADMM emerges as a promising approach for high-dimensional structure learning for physical processes.}

The rest of the paper is organized as follows. We elaborate our derivation of ACLIME-ADMM algorithm in Section 2, along with the stability analysis for hyperparameters. In section 3, PC stable algorithm and how structure learning algorithm is applied for temporal models are illustrated. We provide the description of both synthetic and observed data sets for climate application and the corresponding experimental results in section 4 and section 5 respectively. The advantages of fast implementation of the proposed algorithm is illuminated in section 6 and the paper is concluded in section 7. 

\section{Derivation of ACLIME-ADMM}
\label{sec:related}

Over the past decade, advances in structure learning have been made by making explicit assumptions about the parametric form of the joint distribution. For example, advances have been made based on the assumption that the joint distribution is a multivariate Gaussian \cite{friedman2008sparse,meinshausen2006high,cai2011constrained}, or a Gaussian copula distribution \cite{liu2012high,xue2012regularized}.
Typically, such estimators involve a sparsity inducing optimization problem, and efficient algorithms for solving such problems have been developed \cite{banerjee2008model, hsieh2011sparse}.
In recent work, the CLIME estimator \cite{cai2011constrained} was proposed to estimate  sparse inverse of covariance matrix (precision matrix), which reveals the dependency structure for multivariate Gaussian distribution \cite{lauritzen1996graphical}. For a $p$-dimensional problem, CLIME estimates the sparse precision matrix $\hat{\Omega} \in \R^{p \times p}$ by solving the following linear program (LP):
\begin{equation} \label{eq:clime}
\hat{\Omega} = \underset{\Omega \in \R^{p \times p}}{\argmin} \|\Omega\|_1 \quad \textrm{s.t.} \quad \|C\Omega - I \|_{\infty} \le \lambda~,
\end{equation}
where $\lambda > 0$ is a tuning parameter. Recent work has developed scalable optimization algorithms for the problem, which have been shown to scale to a million dimensions \cite{wang2013large}. In spite of its scalability, the empirical performance of the CLIME estimator is sensitive to the choice of the tuning parameter $\lambda$, and it is usually difficult to make the choice in a rigorous data driven manner~\cite{cai2011constrained, wang2013large}. In recent work, a more powerful adaptive version of CLIME, called ACLIME, has been proposed \cite{cai2016estimating}. In this section, we propose the ACLIME-ADMM algorithm, which is able to solve the corresponding optimization efficiently using block parallel updates along with simple per column adjustments. The introduced inexact ADMM algorithm, which utilizes closed-form updates for both primal and dual variables, improves the scalability of our method considerably.

\subsection{Adaptive Estimation of Statistical Dependencies - Overview}
\label{sec:aclime}

While estimators such as graphical Lasso \cite{meinshausen2006high, friedman2008sparse} and CLIME \cite{cai2011constrained} effectively use the same (soft/box) threshold parameter $\lambda$, recent work on the Adaptive CLIME \cite{cai2016estimating} estimator advocates using a different threshold parameter $\lambda_{ij}$ for different entries. Such a choice arguably leads to better statistical properties of the estimator \cite{cai2016estimating}. Further, the necessary threshold parameters themselves can be obtained in a data driven manner using a suitable estimator.

\subsection{{\it ACLIME} Estimator}
We start by briefly reviewing the ACLIME estimator, the key optimization problems which need to be solved. The following result \cite{cai2016estimating} motivates the estimator:
\begin{theorem} \label{theo:1}
Let $x_1,\cdots, x_n \sim N_p(\mu^*, C^*)$ with $\log p = O(n^{1/3})$, and let $\Omega^*$ be the corresponding precision matrix. Let $C$ be the unbiased sample estimate of $C^*$ and let $S = (s_{ij})_{1\le i,j \le p} = C \Omega^*  - \I_{p \times p}$. Then
\[
Var(s_{ij}) =
\left\{\begin{matrix}
 n^{-1}(1+ c_{ii}^* \omega_{ii}^*) & \text{for} \quad i=j \\
 n^{-1}c_{ii}^* \omega_{jj}^*& \text{for} \quad i \ne j~,
\end{matrix}\right.
\]
and for all $\delta \ge 2$,
\begin{equation}
\mathbb{P} \left\{ |(C \Omega^* - \I_{p \times p})_{i,j}| \le \delta \sqrt{\frac{c_{ii}^* \omega_{jj}^* \log p}{n}}, \forall 1 \le i,j \le p \right\}
\ge 1- O\left((\log p)^{-\frac{1}{2}} p^{ - \frac{\delta^2}{4} + 1} \right)~.\label{eq:bnd1}
\end{equation}
\end{theorem}
To use the adaptive bound in \eqref{eq:bnd1}, one can use the sample estimate $c_{ii}$ as a surrogate to $c^*_{ii}$. However, the bound also needs an estimate of $\omega_{jj}^*$, the diagonal estimates of the precision matrix. The ACLIME estimator works in two stages: in the first stage, an estimate $\breve{\omega}_{jj}$ for $\omega_{jj}^*$ is computed; in the second stage, the estimate $\breve{\omega}_{jj}$ is used to adaptively estimate $\Omega$ based on \eqref{eq:bnd1}. In particular, {\em in the first stage}, each column of the precision matrix is estimated \cite{cai2016estimating} by solving:
\begin{equation} \label{eq:aclime-step1}
\hat{\omega}^1_{.j} = \underset{\bb_j \in \mathbb{R}^p}{\argmin} \{ \|\bb_j\|_1:  
|\hat{C} \bb_j -\e_j|_\infty \le \tau_n (c_{ii} \vee c_{jj}) \times b_{jj}, b_{jj} > 0 \}~,
\end{equation}
where $\hat{C} = C + \frac{1}{n}\I_{p \times p}$, $\tau_n = \delta \sqrt{\frac{\log p}{n}}$, $\delta \geq 2$, $(c_{ii} \vee c_{jj}) = \max(c_{ii},c_{jj})$ and $b_{jj}$ is the $j$-th element in $\bb_j$. Then, the diagonal elements $\omega^*_{jj}$ are estimated as:
\begin{equation} \label{change-diag}
\breve{\omega}_{jj} = \hat{\omega}^1_{jj} I \left\{ c_{jj} \le \sqrt{\frac{n}{\log p}} \right\} + \sqrt{\frac{\log p}{n}} I \left\{ c_{jj} > \sqrt[]{\frac{n}{\log p}}\right\}~.
\end{equation}

Given $\breve{\omega}_{jj}$, {\em in the second stage}, ACLIME estimates $\Omega^*$ by first solving the following optimization problem to get a primitive estimate of the $j$-th column:
\begin{equation} \label{eq:aclime-step2}
\begin{split}
\tilde{\omega}^1_{.j} = & \underset{\bb_j \in \mathbb{R}^p}{\argmin} \{ \|\bb_j\|_1:
 |(\hat{C} \bb_j - \e_j)_i| \le \tau_n \sqrt{c_{ii} \breve{\omega}_{jj}}   \}~.
\end{split}
\end{equation}
In the final step, ACLIME symmetrizes $\tilde{\Omega}^1 = (\tilde{\omega}^1_{ij})$ to obtain $\hat{\Omega} = (\hat{\omega}_{ij})$, the estimate of $\Omega^*$:
\begin{equation} \label{symmetrize-prec}
\hat{\omega}_{ij} = \hat{\omega}_{ji}  = \tilde{\omega}^1_{ij} I \{|\tilde{\omega}^1_{ij}| \le |\tilde{\omega}^1_{ji}|\} + \tilde{\omega}^1_{ji} I \{|\tilde{\omega}^1_{ij}| > |\tilde{\omega}^1_{ji}|\}~.
\end{equation}

\subsection{ {\it ACLIME-ADMM} Algorithm}


We now focus on developing efficient optimization algorithms for solving the two stages of the ACLIME estimation, in particular the problems in \eqref{eq:aclime-step1} and \eqref{eq:aclime-step2}. \cite{cai2016estimating} observes that the optimization problem can be decomposed into $p$ independent LPs, one for each column of $\hat{\Omega}$. We first introduce an inexact ADMM algorithm for solving the column-specific LPs corresponding to each stage, where all computations are in closed form based on elementwise operations and matrix multiplications. Later we generalize the algorithm to solve column block LPs where the computations need more care since the LP for each column is mildly different but has some shared structure which our algorithm uses. 
As the experiments illustrate, the methods are efficient and scalable.
\vspace*{0.3cm}

\noindent {\bf Stage 1: Estimating diagonal elements $\omega_{jj}$.}
We first focus on developing an approach to solving (\ref{eq:aclime-step1}), which yields the initial estimates
of the diagonal elements $\omega_{jj}$ of the precision matrix. We z-score the variables so that $c_{jj} = 1$ for $j=1,\ldots,p$. As a result, considering the constraint in (\ref{eq:aclime-step1}), we note that $\tau_n (c_{ii} \vee c_{jj}) = \tau_n$. Hence the constraint in (\ref{eq:aclime-step1}) can be rewritten as:
\begin{equation} \label{eq:const-aclime1}
-\tau_n b_{jj} \mathbf{1}_{p} \le \hat{C} \bb_j -
\e_j \le \tau_n b_{jj} \mathbf{1}_{p}~,
\end{equation}
where $\mathbf{1}_{p}$ is the $p$ dimensional vector with all entries being 1. Focusing on the right hand side inequality in \eqref{eq:const-aclime1}, we can rewrite it as:
\begin{equation}
\hat{C}_{\up} \bb_j \leq \e_j~,
\label{eq:adm1-up}
\end{equation}
where $\hat{C}_{\up} = \hat{C} - \tau_n \1_p \e_j^T.$
Note that $\hat{C}_{\up}$ is a rank-1 and sparse perturbation of $\hat{C}$ where only column $j$, interacting with $b_{jj}$, gets a constant $\tau_n$ subtracted from every entry. Introducing non-negative variables $\u_j \in \R_+^p$, so that $\u_j \geq \0_p$, the $p$ dimensional vector with all entries being 0, the inequality constraint in \eqref{eq:adm1-up} can be rewritten as an equality constraint:
\begin{equation}
\hat{C}_{\up} \bb_j + \u_j = \e_j~.
\label{eq:adm1-up2}
\end{equation}
Similarly, focusing on the left hand side inequality of \eqref{eq:const-aclime1}, we get
\begin{equation}
-\hat{C}_{\down} \bb_j \leq - \e_j
\label{eq:adm1-down}
\end{equation}
where $\hat{C}_{\down} = \hat{C} + \tau_n \1_p \e_j^T$. Introducing non-negative variables $\v_j \in \R_+^p$, so that $\v_j \geq \0_p$, the inequality constraint in \eqref{eq:adm1-down} can be rewritten as an equality constraint:
\begin{equation}
-\hat{C}_{\down} \bb_j + \v_j = -\e_j~,
\label{eq:adm1-down2}
\end{equation}
Then, by combining  (\ref{eq:adm1-up2}) and \eqref{eq:adm1-down2}, the constraint corresponding to \eqref{eq:const-aclime1} can be written as:
\begin{equation} \label{eq:equality1}
\underbrace{
\begin{bmatrix}
\hat{C}_{\up}\\
-\hat{C}_{\down}
\end{bmatrix}}_{A_j} \bb_j +
\underbrace{\begin{bmatrix}
\I_{p\times p} & 0\\
0 & \I_{p \times p}
\end{bmatrix}}_B\underbrace{\begin{bmatrix}
\u_j\\
\v_j
\end{bmatrix}}_{\r_j}
 =
 \underbrace{\begin{bmatrix}
 \e_j\\
 -\e_j
 \end{bmatrix}}_{\c_j}~.
 \end{equation}
Then, the original problem in \eqref{eq:aclime-step1} can be written in a canonical form suitable for ADMM as follows:
\begin{equation}
\min_{\bb_j \in \R^p,\r_j \in \R^{2p}}~ \| \bb_j \|_1 + \ind_{\R_+}(\r_j) \quad \text{s.t.} \quad A_j \bb_j + \r_j = \c_j~,
\label{eq:aclime-admm1}
\end{equation}
where $\ind_{\R_+^{2p}}(\cdot)$ is the indictor function over non-negative reals in $\R^{2p}$, i.e.,
$\ind_{\R_+^{2p}}(\z_j) = 0$, if $\z_j \ge \0_{2p}$, and $\infty$ otherwise, and we have used the fact $B = \I_{2p \times 2p}$, the identity matrix.

The augmented Lagrangian of the optimization problem in (\ref{eq:aclime-admm1}) is :
\begin{equation} \label{eq:lag-1}
L(\bb_j,\r_j,\y_j)= \|\bb_j\|_1 + \ind_{\R_+^{2p}}(\r_j) + \rho \langle \y_j, A_j\bb_j+\r_j-\c_j\rangle + \frac{\rho}{2} \|A_j \bb_j+\r_j-\c_j\|_2^2~,
\end{equation}
where $\y_j \in \R^{2p}$ is the Lagrange multiplier vector.
Based on the augmented Lagrangian, the ADMM steps are:
\begin{subequations} \label{eq:admm-updates}
\begin{alignat}{3}
\bb_j^{t+1} &= \underset{\bb_j \in \R^p}{\argmin} \|\bb_j\|_1 + \frac{\rho}{2} \|A_j\bb_j + \r_j^t-\c_j + \y_j^t\|_2^2 \label{eq:primal1} \\
\r_j^{t+1} &= \underset{\r_j \in \mathbb{R}^{2p}}{\argmin} \ind_{\R_+^{2p}}(\r_j) +\frac{\rho}{2} \| A_j \bb_j^{t+1}+ \r_j-\c_j+ \y_j^t\|_2^2 \label{eq:primal2}\\
\y_j^{t+1} &= \y_j^t + A_j\bb_j^{t+1} + \r_j^{t+1}-\c_j~. \label{eq:dual1}
\end{alignat}
\end{subequations}
The update of $\bb_j$ in \eqref{eq:primal1} does not have a closed form solution because the $A_j^T A_j$ term makes the components of $\bb_j$ coupled. While one can use iterative approaches to solve the problem, we decouple the $\bb_j$ by linearizing the quadratic term and adding a proximal term, a strategy used in inexact ADMM \cite{boyd2011distributed}:
\begin{equation}\label{eq:in-admm-bj}
\bb_j^{t+1} = \arg \min_{\bb_j \in \R^p} \|\bb_j\|_1 + \eta \langle \g_j^t, \bb_j\rangle + \frac{\eta}{2} \|\bb_j-\bb_j^t\|_2^2~,
\end{equation}
where $\g_j^t = \frac{\rho}{\eta} A_j^T(A_j\bb_j^t+\r_j^t - \c_j + \y_j^t)$ and $\eta > 0 $. Inexact ADMM has been shown to have the same rate of convergence as ADMM for general (non-smooth) convex optimization problems \cite{wang2014bregman}.
Now, based on the dual update in \eqref{eq:dual1}, we have $\g_j^t = \frac{\rho}{\eta} A_j^T(2\y_j^t - \y_j^{t-1})$.
Then, (\ref{eq:in-admm-bj}) has the following closed form solution based on soft-thresholding \cite{boyd2011distributed}
\begin{equation}
\label{eq:soft}
\bb_j^{t+1} = \text{soft}(\bb_j^t - \g_j^t, \frac{1}{\eta})~.
\end{equation}
Updating $\r_j^{t+1}$ in \eqref{eq:primal2} is simply the projection of elements of $\h_j^t = \c_j- \y_j^t - A_j\bb_j^{t+1}$ to $\R_+^{2p}$ which can be done in closed form as $\r_j^{t+1} = \max(\h_j^t,0)$, applied elementwise. 


The solution of the above optimization for stage 1 gives $\hat{\omega}^1_{\cdot j}$ in (\ref{eq:aclime-step1}), from which only the diagonal elements $\hat{\omega}^1_{jj}$ are of interest, which are then used to compute $\breve{\omega}_{jj}$ following (\ref{change-diag}).

\vspace*{0.3cm}

\noindent {\bf Stage 2: Estimating $\Omega$.}
In the second stage of ACLIME, the goal is to utilize the $\breve{\omega}_{jj}$ estimated in stage 1, and solve the problem in (\ref{eq:aclime-step2}) to obtain $\tilde{\omega}_{\cdot j}$. Considering the constraints in (\ref{eq:aclime-step2}), since $c_{ii} = 1$ due to z-scoring, the constraints over $\bb_j \in \R^p$ can be simplified to
\begin{equation}
- \tau_n \sqrt{\breve{\omega}_{jj}} \1_p \leq \hat{C} \bb_j - \e_j \leq \tau_n \sqrt{\breve{\omega}_{jj}}
\end{equation}
Then, following the same strategy as used for stage 1, the system of linear inequality constraints can be rewritten as a system of equality constraints
\begin{equation} \label{eq:equality1_stage2}
\underbrace{
\begin{bmatrix}
\hat{C}\\
-\hat{C}
\end{bmatrix}}_{A} \bb_j +
\underbrace{\begin{bmatrix}
I_{p\times p} & 0\\
0 & I_{p \times p}
\end{bmatrix}}_B\underbrace{\begin{bmatrix}
\u_j\\
\v_j
\end{bmatrix}}_{\r_j}
 =
 \underbrace{\begin{bmatrix}
 \e_j + \tau_n \sqrt{\breve{\omega}_{jj}} \1_p \\
 -\e_j + \tau_n \sqrt{\breve{\omega}_{jj}} \1_p
 \end{bmatrix}}_{\c_j}~.
 \end{equation}
where $\r_j \in \R_+^{2p}$ as before.
Then, the original problem in \eqref{eq:aclime-step2} can be written in a canonical form suitable for ADMM as follows:
\begin{equation} 
\min_{\bb_j \in \R^p,\z_j \in \R^{2p}}~ \| \bb_j \|_1 + \ind_{\R_+}(\z_j) \quad \text{s.t.} \quad A \bb_j + \z_j = \c_j~.
\label{eq:aclime-admm2}
\end{equation}
We note that the optimization problem in (\ref{eq:aclime-admm1}) is essentially the same as that in (\ref{eq:aclime-admm2}), in fact simpler since $A$ is the same for all $j$. One can use the same ADMM algorithm for stage 2, take advantage of the same structures in the matrices to speed up computations, and also perform block updates which are going to be simpler since $A$ is the same for all $j$.

Given that the structure of the optimization in stage 2 is simpler, one can also consider an alternative route \cite{wang2013large}, which uses less variables and is arguably amenable to block updates. Note that since $c_{ii} = 1$ due to z-scoring, the problem in  (\ref{eq:aclime-step2}) can be posed as:
\begin{equation}
\min_{\bb_j \in \R^p}~\| \bb_j \|_1 \quad \text{s.t.} \quad \| \hat{C} \bb_j - \e_j \|_{\infty} \leq \lambda_j \end{equation}
where $\lambda_j = \tau_n \sqrt{\breve{\omega}_{jj}}$ is a constant. Introducing $\z_j \in \R^p$, the problem can be rewritten as
\begin{equation}
\min_{\bb_j,\z_j \in \R^p}~\| \bb_j \|_1 \quad \text{s.t.} \quad \| \z_j - \e_j \|_{\infty} \leq \lambda_j~, \hat{C} \bb_j = \z_j~.
\end{equation}
Note that the constraint on $\z_j$ is a box constraint, on which efficient projection is possible. Hence the box constraint can be handled inside the primal update for $\z_j$, without having to convert the box constraint to a system of equality constraints. Thus, ignoring the box constraint for now, the augmented Lagrangian is
\begin{equation}
L(\bb_j,\r_j,\y_j) = \| \bb_j \|_1 + \rho \langle \y_j, \hat{C}\bb_j - \z_j \rangle + \frac{\rho}{2} \| \hat{C}\bb_j - \z_j \|^2~.
\end{equation}
The ADMM updates, which take the box constraint into account, are as follows
\begin{subequations} \label{eq:admm-updates2}
\begin{alignat}{3}
\bb_j^{t+1}& = \underset{\bb_j \in \R^p}{\argmin} \| \bb_j \|_1 + \frac{\rho}{2} \| \hat{C}\bb_j - \z_j^t + \y_j^t\|^2~ \label{eq:primal3} \\
\z_j^{t+1} & = \underset{\| \z_j - \e_j \|_{\infty} \leq \lambda_j}{\argmin} \frac{\rho}{2} \| \hat{C}\bb_j^{t+1} - \z_j + \y_j^t \|^2 \label{eq:primal4}\\
\y_j^{t+1} & = \y_j^t + \hat{C} \bb_j^{t+1} - \z_j^{t+1}~. \label{eq:dual2}
\end{alignat}
\end{subequations}
Note that \eqref{eq:primal3} can be solved using an inexact update similar to \eqref{eq:in-admm-bj}. Further, we note that the box-constrained quadratic problem in \eqref{eq:primal4} can be solved in closed form as
\begin{equation}
\z_j^{t+1} = \text{box}(\hat{C}\bb_j^{t+1} + \y_j^t,\e_j,\lambda_j)
\label{eq:p3update}
\end{equation}
where for $\a,\w \in \R^p, \lambda \in \R_+$
\begin{equation}
\text{box}(\a,\w,\lambda) = \begin{cases}
w_i + \lambda~,  & \quad \text{if}~~a_i - w_i > \lambda \\
a_i~,  & \quad \text{if}~~|a_i - w_i| \leq \lambda \\
w_i - \lambda~,  & \quad \text{if}~~a_i - w_i < -\lambda ~.
\end{cases}
\end{equation}
In the current setting, $\a = \hat{C}\bb_j^{t+1} + \y_j^t, \w = \e_j$, and $\lambda = \lambda_j$.


The solution of the above optimization in stage 2 gives $\tilde{\omega}_{\cdot j}$ in (\ref{eq:aclime-step2}). The final step is to symmetrize the resulting precision matrix estimate as in \eqref{symmetrize-prec}.

\subsection{Column-Block {\it ACLIME-ADMM} Algorithm}

We propose an improvement to solve the two-stage ACLIME optimization in terms of column blocks instead of column-by-column.The implementation for each step is either element-wise parallel or utilizes suitable matrix multiplication, which improved the computational efficiency of the proposed algorithm. For stage one, we rewrite $A_j\bb_j$ as following:
\begin{equation}\label{eq:revise}
\begin{split}
\MoveEqLeft
A_j\bb_j=
\begin{bmatrix}
\hat{C}_{\text{up}}\bb_j \\
\hat{C}_{\text{down}}\bb_j
\end{bmatrix}
=
\begin{bmatrix}
\hat{C}x-\tau_n b_{jj} \1_p\\
-\hat{C}x-\tau_n b_{jj} \1_p\\
\end{bmatrix}\\
&
=
\begin{bmatrix}
\hat{C} \\
-\hat{C}
\end{bmatrix}
\bb_j-\tau_n b_{jj} \1_{2p}~.
\end{split}
\end{equation}
Since all $A_j$ are transformed from $\hat{C}$, the computation across columns can be shared, e.g., computing $\hat{C} \bb_j$. Now we consider the column blocks, assuming $X \in \mathbb{R}^{p \times k}$ denotes $k$ columns in $\hat{\Omega}$. Thus, the $AX$ for a column block is defined as:
\begin{equation}\label{eq:revise15}
AX=
\begin{bmatrix}
\hat{C}\\
-\hat{C}
\end{bmatrix}
X
-\1_{2p \times k} X_{\text{diag}}~,
\end{equation}
where $X_{\text{diag}}\in \mathbb{R}^{k \times k}$ is a diagonal matrix with the corresponding diagonal elements in $X$ and $\1_{2p \times k}\in \mathbb{R}^{2p\times k}$ is a matrix with all entries being 1. Therefore, the equality constraints (\ref{eq:equality1}) for column block is $AX+R=E$, where $R$ is the column block of corresponding $\r_j$ in (\ref{eq:aclime-admm1}) and $E\in \mathbb{R}^{p \times k}$ denotes the same $k$ columns in $\I_{p\times p}$. 

\begin{figure}[ht]
\subcaptionbox{Undirected Graph}%
  [.3\textwidth]{\includegraphics[scale=0.40]{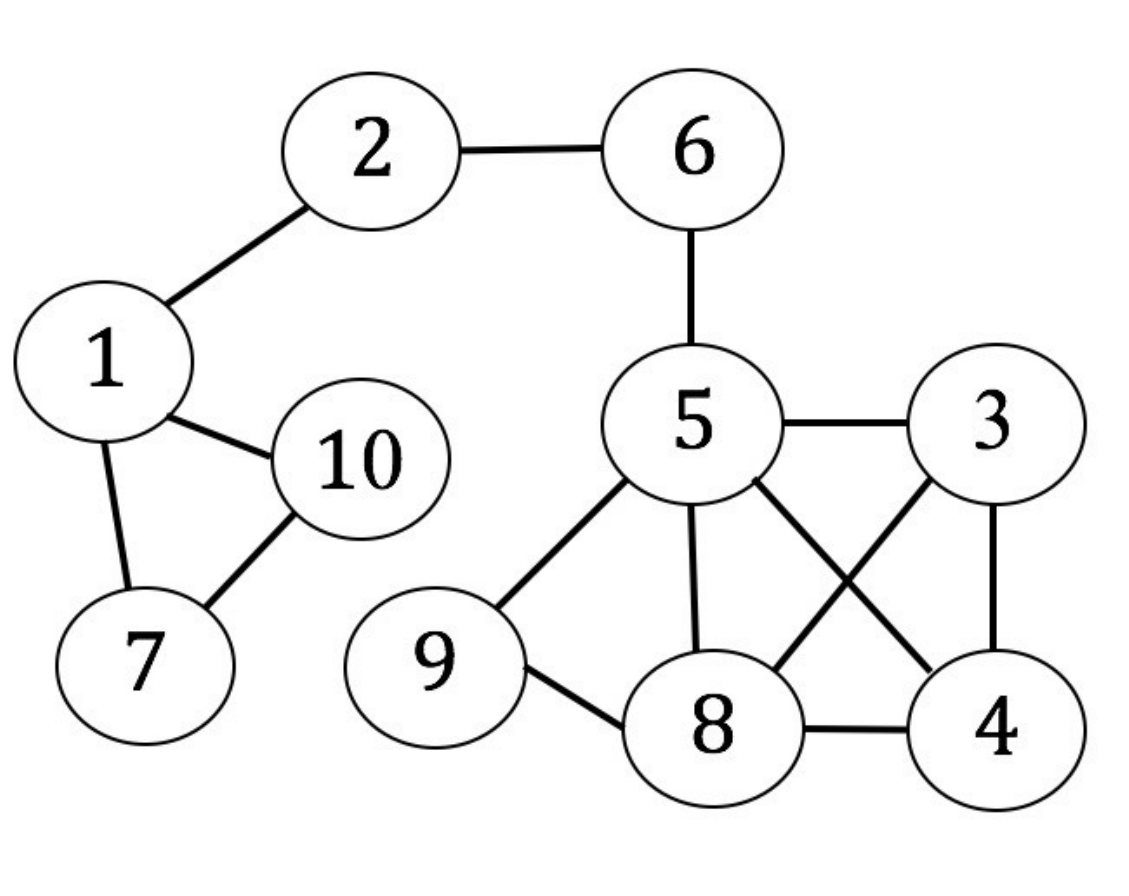}  \vspace{3mm}}
  \subcaptionbox{Residuals}%
  [.3\textwidth]{\includegraphics[scale=0.25]{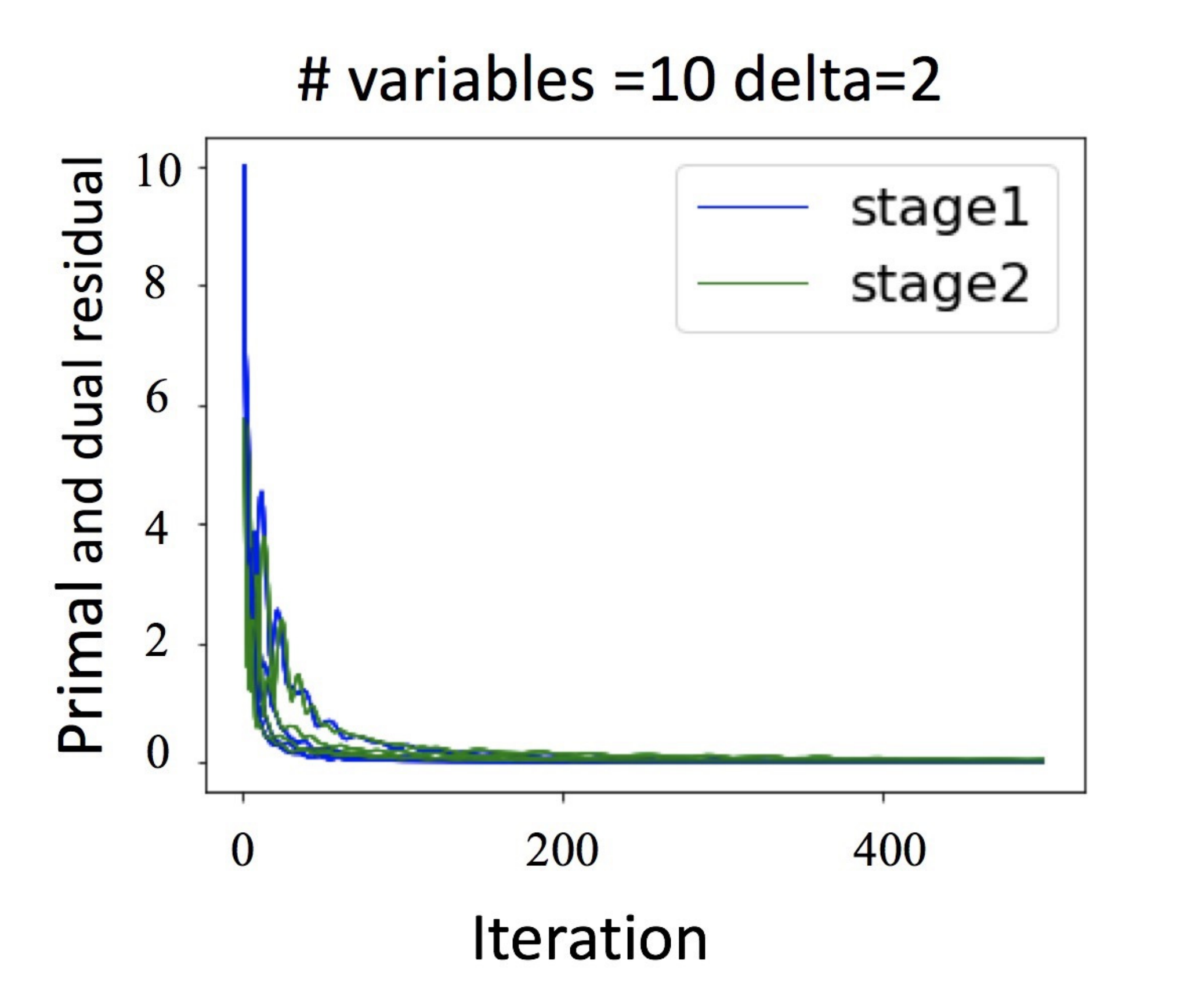}}
  \subcaptionbox{Accuracy}%
  [.3\textwidth]{\includegraphics[scale=0.15]{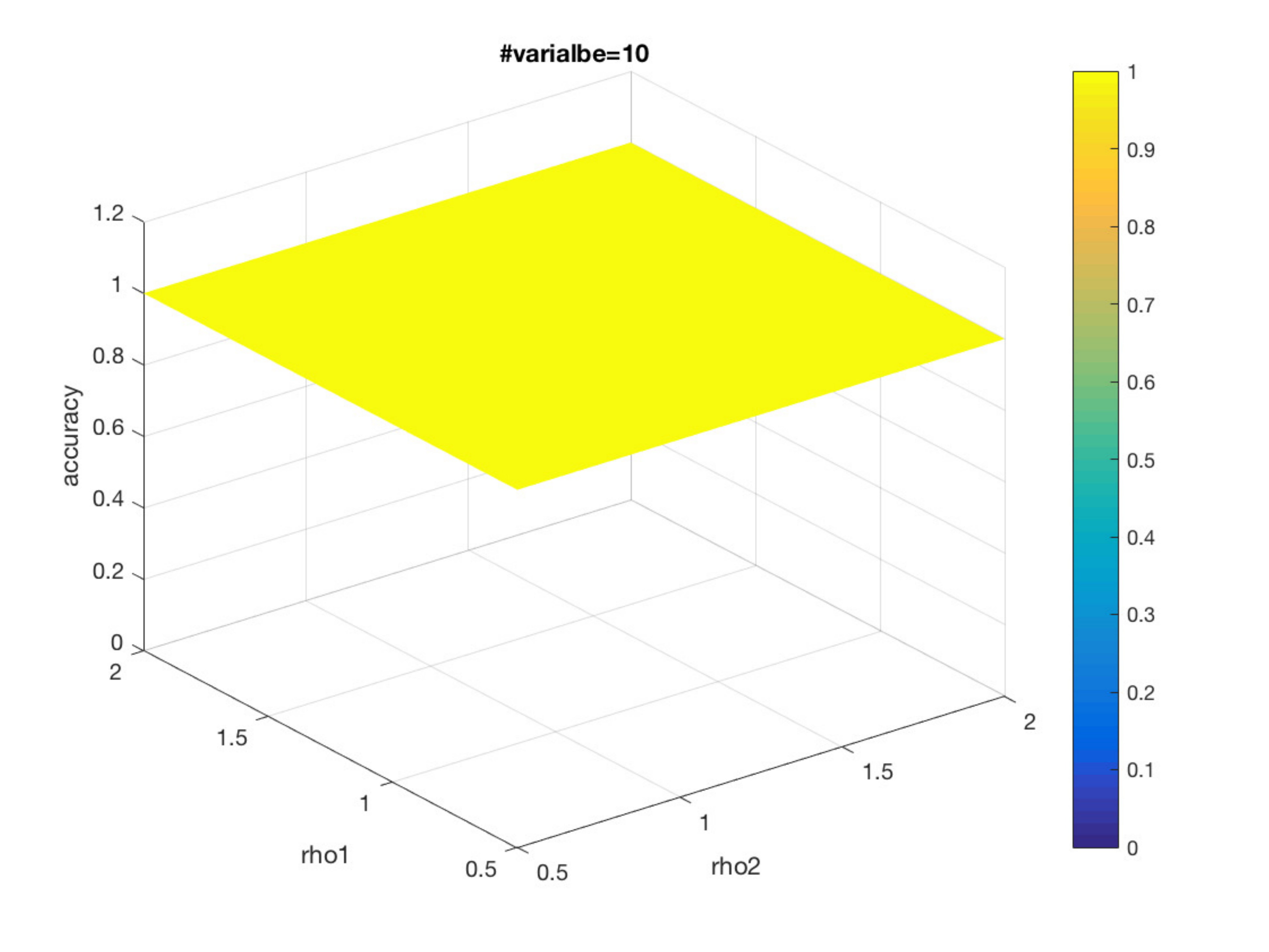}}
\caption{The results of 10-variable synthetic data from column-block ACLIME-ADMM. (a) It is the true underlying undirected graph for the synthetic data, which illustrates the dependencies among 10 variables. (b) The primal and dual residuals of two stages in column-block ACLIME-ADMM are shown converging to 0 after 400 iterations when $\delta=2$. (c) When parameter $\rho$ in two stages are chosen in $(0,2]$, the accuracy for estimating the non-zero entries in precision matrix are always 100\% for the synthetic data.}
\label{fig:stability}
\end{figure}

Therefore, the optimization problem is rewritten as follows:
\begin{equation}
\begin{matrix}
\underset{X\in \mathbb{R}^{p \times k}, R \in \mathbb{R}^{2p \times k}} {\min}\norm{X}_1 + \mathbbm{1}_{\mathbb{R}^{2p \times k}_+}(R) \\
\text{s.t.} \quad AX+R=Z~.
\end{matrix}
\end{equation}
Thus, the augmented Lagrangian of the above optimization problem is 
\begin{equation}
L_\rho = \norm{X}_1 + \mathbbm{1}_{\mathbb{R}^{2p \times k}_+}(R) +\rho \langle Y, AX+R-E \rangle+\frac{\rho}{2} \norm{AX-R-E}^2_2,
\end{equation}
where $Y \in \mathbb{R}^{p \times k}$ is a scaled dual variable and $\rho>0$. Similar to (\ref{eq:admm-updates2}), inexact ADMM yields the following iterates:
\begin{subequations}
\begin{alignat}{3}
X^{t+1}& =\underset{X\in \mathbb{R}^{p \times k}}{\argmin}  \norm{X}_1+\eta \langle V^t , X \rangle + \frac{\eta}{2} \norm{X-X^t}^2_2\label{eq:block-ADMM1}\\
R^{t+1}& =\underset{R \in \mathbb{R}^{2p \times k}}{\argmin}  \mathbbm{1}_{\mathbb{R}^{2p \times k}_+}(R)+\frac{\rho}{2}\norm{AX^{t+1}+R-E+Y^t}_2^2\\
Y^{t+1}& =Y^{t}+AX^{t+1}+R^{t+1}-E~,\label{eq:block-ADMM3}
\end{alignat}
\end{subequations}
where $V^t=\frac{\rho}{\eta}A^T(2Y^t-Y^{t-1})$. Then (\ref{eq:block-ADMM1}) has a closed form solution based on element-wise soft-thresholding
$X^{t+1}=\text{soft}(X^t-V^t, \frac{1}{\eta})$.
The only problem left is how to compute $A^TY$ in $V^t$, which can be solved as
 \begin{equation}\label{eq:AY}
A^TY=
\begin{bmatrix}
\hat{C} & -\hat{C}
\end{bmatrix}
\begin{bmatrix}
Y_1 \\
Y_2
\end{bmatrix}
-W_{\text{diag}}~,
\end{equation}
 where $Y_1, Y_2 \in \R^{p\times k}$ are respectively upper and lower half of $Y$
 and $W_{\text{diag}}=W_{\text{diag1}}-W_{\text{diag2}}.$ Assume the $k$-column block matrix $X$ contains $(i+1)$-th column to $(i+k)$-th column in $\hat{B}$, then $W_{\text{diag1}}$ is
 \begin{equation}
     W_{\text{diag1}}=
     \begin{bmatrix}
     \mathbf{0_1}\\
     \mathbf{D}_{k\times k}\\
     \mathbf{0_2}
     \end{bmatrix}~,
 \end{equation}
 where $\mathbf{0_1}\in \R^{i\times k}$ and $\mathbf{0_2}\in \R^{(p-i-k)\times k}$ are matrices with all zero entries. $\mathbf{D}_{k \times k}\in \R ^{k \times k}$ is a diagonal matrix block, in which the $m$-th diagonal element $d_{m,m}=\tau_n {Y_1}_{(m+1)}^T \1_p$ and ${Y_1}_{(m)}$ is the $m$-th column of $Y_1$ in (\ref{eq:AY}). $W_{\text{diag2}}$ has the same format based on $Y_2$. The update of $R^{t+1}$ can be done in closed form as $R^{t+1}=\max(H^t,0)$, applied elementwise, where $H^t=E-Y^t-AX^{t+1}$.

For stage two, the problems for different columns only differ in the threshold $\lambda_j$ in \eqref{eq:primal3}, therefore the corresponding update in \eqref{eq:p3update} can be done in element-wise parallel manner for column blocks.

\textbf{Stability of Column-Block ACLIME-ADMM}
 The inexact ADMM algorithm introduced two parameters, i.e., the scaled stepsize $\rho$ and linearization parameter $\eta$. In \cite{wang2014bregman}, it is proved that the value of $\eta$ depends on the convexity of the objective function. The experimental results for synthetic datasets show that the proposed algorithm is stable within the reasonable range of $\rho$. For a fixed $\eta$, the converge rate can be guaranteed if $\eta \ge \rho \lambda^2_{max}(C)$, where $\lambda_{max}(C)$ is the largest eigenvalue of covariance matrix. We validate the stability with a 10-variable synthetic dataset with 1500 samples, in which the variables follow multivariate Gaussian distribution. The underlying undirected graph is shown in Fig.\ \ref{fig:stability}(a). Fig.\ \ref{fig:stability}(b) and Fig.\ \ref{fig:stability}(c) show that the primal and dual residual converges to 0 and the estimated matrix can always detect the non-zero elements (i.e., the undirected edge in graph) correctly when $\rho$ for both stages are chosen in $(0,2]$. The proposed column-block ACLIME-ADMM can be achieved based on the parallel processing for separate column blocks, which leads to the high-efficiency and scalability. The estimation of precision matrix for large scale datasets is solvable with limited working memory.



\section{PC stable and Temporal Models}

We use a variation of the classic {\it PC} algorithm as baseline algorithm for comparison. This section provides details of that algorithm and explains how structure learning algorithms can be used to derive temporal models.

\subsection{PC stable algorithm}

One of the best-known algorithms for structure learning is the well-known
{\it PC} algorithm \cite{spirtes1991algorithm}. Colombo and Maathuis \cite{colombo2014order}  developed an improved version of the PC algorithm, called PC stable. PC stable is order-independent, more robust and easy to parallelize, and is used in this paper.  PC stable has only one parameter to choose, the significance value $\alpha$ for the statistical independence tests. We used $\alpha = 0.05$ for the runs with synthetic data and $\alpha = 0.1$ for the runs with observed data. There is generally little difference in the output of the PC stable algorithm for varying values of $\alpha$ (even up to $\alpha = 0.5$), so such a small change has no relevance for the results.


\subsection{From Static to Temporal Model}

Structure learning methods, including PC stable, CLIME-ADMM and ACLIME-ADMM,
treat their input data as static data, i.e.\ the order of the
samples does not matter.
Most data in the geosciences, however, comes from temporal processes and
the order of, and temporal distance between, samples is crucial for their interpretation.
We can adapt structure learning algorithms to incorporate that information and
to capture those temporal relationships explicitly
using the approach first proposed by Chu et al.\ \cite{chu2005data}.
The key idea is to introduce lagged variables
into the model that capture the relationship between variables at different
instances in time.
The data of those lagged variables is populated from the original data and
encapsulates the temporal information.
In effect, we can thus turn a data set with $q$ variables and temporal information
into a data set with $p =(q \times T)$ variables, where $T$ is the number of lagged copies for each variable. The new dataset can be treated as a static data set,
and thus can be handled by standard structure learning algorithms.
Once the static model with lagged variables is solved,
the output can be converted to model the original variable set but complete with
temporal relationships.
The price to pay for this temporal model is high complexity, because
rather than dealing with $q$ variables, we deal with $p = (q \times T)$.
This is another reason why we often encounter very high-dimensional
problems in the geosciences. There are some associated initialization issues, but those can easily be overcome \cite{ebert2014causal}.
We adopt this approach for all algorithms used here. For the synthetic datasets (see Section~\ref{sec:results}), we have $q = 400, T=20$, so that $p = 8,000$ with $n = 5,200$ samples; for the real dataset, $q=800, T=15$, so $p=12,000$ with $n=4,500$ samples. Note that since {\it ACLIME-ADMM} works with $p^2$ edges in each stage, the optimization for synthetic data involves 64 million variables and that for the real data involves 144 million variables.


\section{Synthetic and Observed Data Sets for Climate Applications}
\label{sec:physical}

\begin{figure*}[t]
\begin{subfigure}[t]{.33\textwidth}
    \includegraphics[scale = 0.19,trim=0 0 0 70bp,clip]{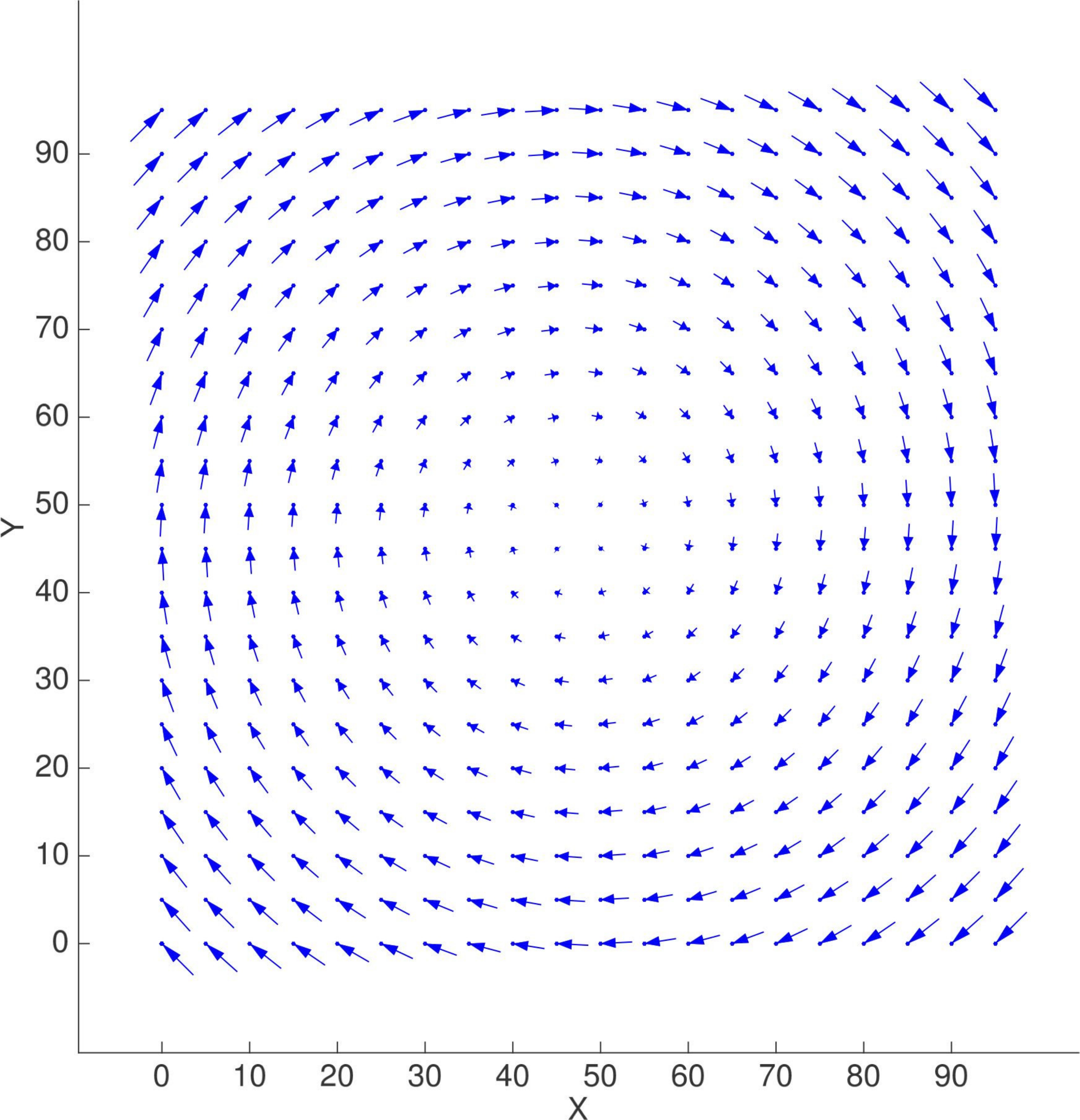}
    \caption{Scenario 1: Circular flow}
\end{subfigure}
\vspace*{3mm}
\begin{subfigure}[t]{.33\textwidth}
     \includegraphics[scale = 0.19,trim=0 0 0 70bp,clip]{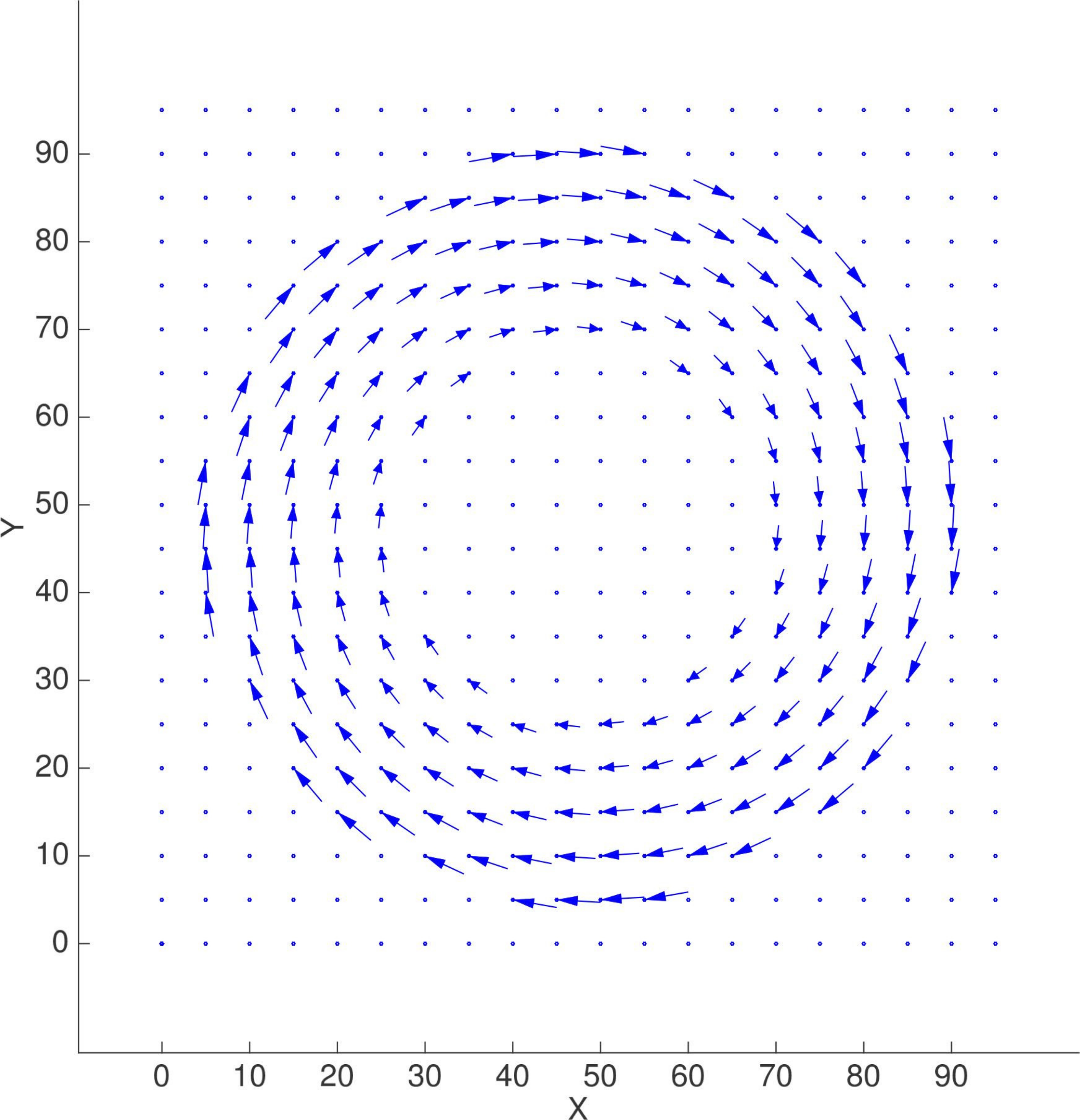}
    \caption{Scenario 2: Ring flow}
\end{subfigure}
\vspace*{3mm}
\begin{subfigure}[t]{.33\textwidth}
    \includegraphics[scale = 0.19,trim=0 0 0 70bp,clip]{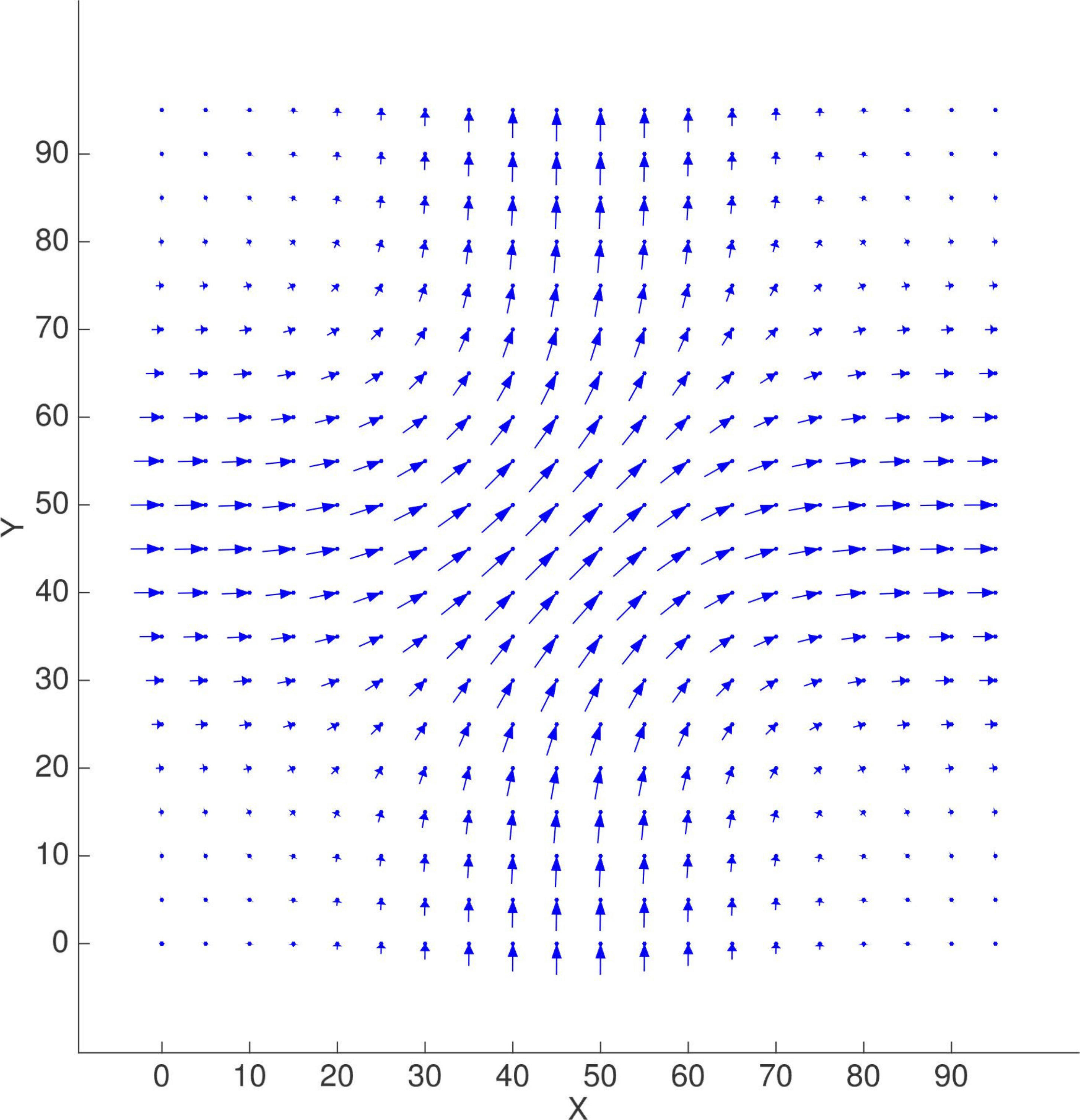}
    \caption{Scenario 3: Cross current}
\end{subfigure}
\vspace*{-5mm}

\caption{Advection Velocity Fields for the three scenarios, which can be interpreted as velocity of fluid flow at each point of square grid.}
\label{scenarios}
\end{figure*}

\subsection{Simulated Advection-Diffusion Processes}

As a testbed for structure learning we created a simulation of a two-dimensional
advection-diffusion process. This testbed generates synthetic data sets with known
diffusion and advection properties,
as benchmarks to test and compare different structure learning algorithms.
We selected advection (e.g. transfer of heat through movement of a fluid) and diffusion (e.g. spread of heat in a resting fluid) processes, because in many geoscience applications
they represent the two most dominant processes.

\remove{\color{red}
Advection describes how a moving fluid - or air - can transport a property or substance.
A classic example is the transport of heat in a moving fluid, for example carrying a warm pocket of water down a stream.
Diffusion describes the spread of a property or substance from a central point
outwards while the fluid stays in place.  A classic example is a pocket of warm
water in a bucket of cold water, where the heat slowly radiates from the warm spot
to colder areas.
}
%
%
The two-dimensional advection-diffusion process is described 
by the following partial differential equation (PDE):
\begin{equation}
    \frac{\partial f}{\partial t} + \left( V_x \frac{\partial f}{\partial x}
   + V_y \frac{\partial f}{\partial y} \right)
= \left( \kappa_x \frac{\partial^2 f}{\partial x^2} + \kappa_y \frac{\partial^2 f}{\partial y^2} \right),
\end{equation}
where $f(x,y,t)$ can be interpreted as the temperature of a fluid at location $(x,y)$
over time $t$,
$\kappa_x$ and $\kappa_y$ are the diffusion coefficients in $x$ and $y$-direction,
respectively, and $V(x,y)$ is the velocity vector field that describes the advection
velocity at any point $(x,y)$.
For the results described here diffusion is symmetric,
$\kappa_x = \kappa_y = \kappa$. We use a square grid
with periodic boundary conditions, i.e.\ we apply a wrap-around in both
$x$ and $y$ direction. To ensure that the connectivity between the grid points is encoded in the data,
we send one signal to each grid point (one at a time) to disturb the system from equilibrium, let the signal travel to other points and dissipate, then repeat the process with the next grid point. Finally, we create three distinct scenarios for testing by choosing three different 
advection fields.
In Scenario 1 (Fig.\ \ref{scenarios}(a)) the advection field is circular
and the magnitude of the velocity is proportional to the distance of the grid point
from the grid center.  Note that the velocity direction near the boundaries is discontinuous because of the wrap-around at the boundaries.
Scenario 1 tests the effect of discontinuity.
In Scenario 2 (Fig.\ \ref{scenarios}(b)) the advection velocity is non-zero only in a ring shape.  Inside and outside of that ring, advection velocities are zero, i.e.,\
in those areas only diffusion is present.  Scenario 2 can thus be used to
test the algorithms for larger areas with only diffusion.
In Scenario 3 (Fig.\ \ref{scenarios}(c)) there are two crossing currents.
One flows from left to right, the other from bottom to top.
Advection velocities outside the main currents are small, but not zero.

\subsection{Observed Data}
\label{ssec:real}
We use data from the NCEP-NCAR reanalysis project \cite{kalnay1996ncep}. The NCEP-NCAR reanalysis project provides data on a global grid for a variety of atmospheric variables and is derived
from observations, but also incorporates the output of numerical weather predictions to improve the quality of the data.
We use daily geopotential height data at 500mb, which denotes for any location the {\it height} at which the air pressure is 500mb.
Data from the years 1950-2000 is used here, and, in order to focus on the dynamics of only one season only daily data from the boreal winter months
(Dec, Jan, Feb) are used.
Since irregularities in the grid, such as varying cell size, are known to create artifacts in the results of structure learning \cite{ebert2014causal}, the data is interpolated on an 800-point grid of nearly equally distributed points, called {\it Fekete} points, on the sphere \cite{bendito2007estimation}.

\section{Experimental Results}
\label{sec:results}
\label{experiment_sec}

In this section we compare results from the {\it PC stable}, {\it CLIME-ADMM} and {\it ACLIME-ADMM} methods for synthetic and real world data. We first discuss the experimental setup and implementation details.


\subsection{Interpretation and Error Measures}

The result of each structure learning algorithm is an adjacency matrix that 
describes which nodes in the graph are connected. 
Since we are learning a temporal model, 
each node in the graph represents a location (grid point) coupled with 
a specific time stamp. 
Thus each connection from the adjacency matrix represents a connection between two physical locations 
along with the two time stamps, 
so we can deduct the time it took to travel from potential source to potential effect. 
Connections with identical time stamps are interpreted as undirected edges.
The remaining edges are directed, going from the location with the earlier time stamp to the one 
with the later time stamp. While it might be tempting to try to develop error measures directly for those
edges (or for the corresponding adjacency matrices), those would be misleading.  
The reason is that 
physical connections do not have a {\it unique} representation in this space.
For example, a signal that travels one grid point in one time step 
can be represented by a connection spanning one grid point distance in one time step,
or by a connection spanning two grid point distances in two time steps, or both.
More generally, 
there are many ways in which signal propagation  
can be represented in this framework, and methods should not be punished for using 
different, legitimate representations.
The way to resolve this problem is {\it to focus on physically meaningful quantities}, 
since those are by definition unique.
In this case a natural choice is to calculate an estimated velocity field, 
i.e.\ for each grid point we estimate a velocity vector by taking the average
of all directed edges incident at the grid point, with each edge normalized by its travel time, $T$,
which is the difference between the time stamps of its two end points. 
(We include both incoming and outgoing edges at each grid point to increase the robustness of the estimates.)
This results in an estimated velocity vector at
each grid point, which then can be compared directly to the advection velocities 
shown in Fig.\ \ref{scenarios}.


Even if the structure learning method was perfect, we could not expect an 
exact match between the two fields\textemdash because of simulation errors and the fact that the advection field does not model the diffusion effects\textemdash but the results should
be very similar to each other.  Thus this is the best ground truth we can get
for such a physical set-up.  

Note that we can provide error measures only for the synthetic data, since the
observed data does not have any quantitative ground truth. For the observed data 
we also generate velocity plots and compare them (visually) to domain knowledge 
in the geosciences.




We use the following error measures.
Numbering the grid points from $i=1$ to $400$,
let $L_i^{\text{adv}}$, $\alpha_i^{\text{adv}}$ denote the length and angle of the advection 
velocity field at point $i$.  
$\hat{L}_i, \hat{\alpha}_i$ denote the 
corresponding velocity estimates obtained through structure learning. 
Then 
$\Delta \alpha_i = \text{abs}( \alpha_i^{\text{adv}} - \hat{\alpha}_i)$ denotes the 
absolute angle error and 
$\Delta L_i = \text{abs}( L_i^{\text{adv}} - \hat{L}_i)$ denotes the 
absolute length error at Point $i$.
Note that if either the advection field or the approximation has zero velocity 
at a grid point, then length $\Delta L_i$ is still well defined, while 
angle $\Delta \alpha_i$ is undefined.  Note that if the velocity is zero 
in {\it both} advection and estimated velocity, we set $\Delta \alpha_i=0$.

We report the following error measures:
\begin{itemize}
\item
   RMSE-Length: The root mean square error of $\Delta L_i$;
\item
   RMSE-Angle: The root mean square error of $\Delta \alpha_i$,
   taking only points into account for which $\Delta \alpha_i$ is well defined.
\item
   PPDL15: The percentage of points for which $\Delta  \alpha_i \! \le \!\! 15$ degrees,
   out of all points for which $\Delta \alpha_i$ is well-defined.
\end{itemize}
Ideally, we want both RMSE measures to be small and the percentage value 
PPDL15 as close as possible to 100.
From a geoscience viewpoint, the direction of connections is generally more important than the exact speed of 
signal travel, thus the angle-related measures are more important than the length-related measures.
To highlight the angle accuracy in the velocity plots for synthetic data,
arrows in these plots are colored based on their angle deviation, $\Delta \alpha_i$.  
The color code is as follows: blue for deviation of $[0,15]$ degrees, black for $(15,30]$ degrees, yellow for $(30,45]$ degrees, and red for $(45, 180]$ degrees.  Furthermore, if the input velocity is zero, and the output velocity is non-zero, then the deviation angle, and thus color, is undefined.  In that case a small length of the output vector indicates a better match, so colors are chosen as follows in that case: blue for length of $[0,0.1]$, black for length of $(0.1,0.5]$, and red for a length of $(0.5, \infty)$.

\remove{
{\color{red}\ab{how do we preprocess the data, how do we threshold the edges, etc. couple of line should be good}.  Imme says: We do not threshold the edges at all. All edges are shown. We threshold edges only for the observed data - which will be discussed there.} 
}


\begin{figure}[ht]

\begin{subfigure}[t]{.3\textwidth}
    \includegraphics[scale = 0.25,trim=0 0 0 110bp,clip]{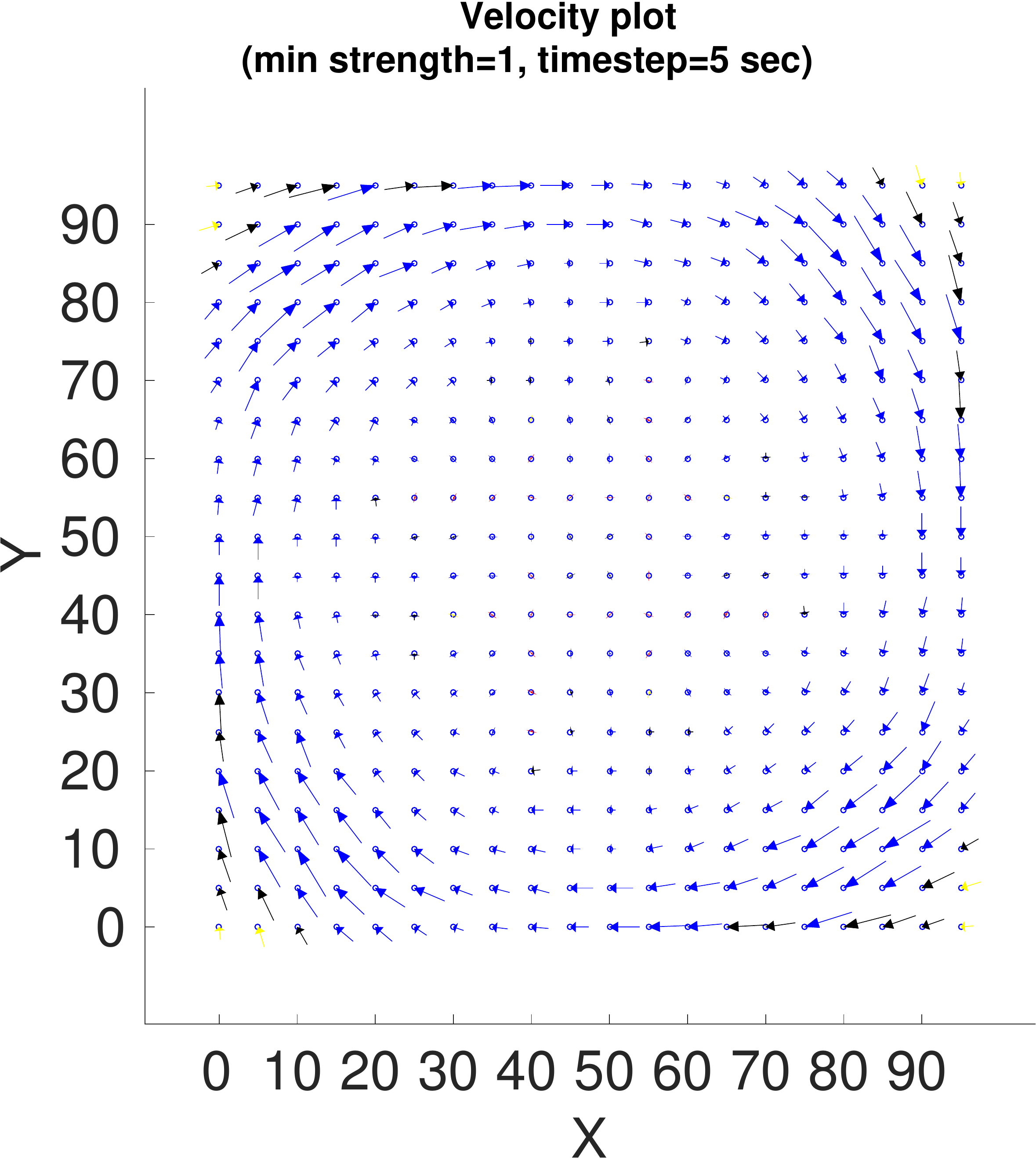}
    \caption{Scenario 1: Circular flow}
\end{subfigure}
\hspace*{3mm}
\begin{subfigure}[t]{.3\textwidth}
    \includegraphics[scale = 0.25,trim=0 0 0 110bp,clip]{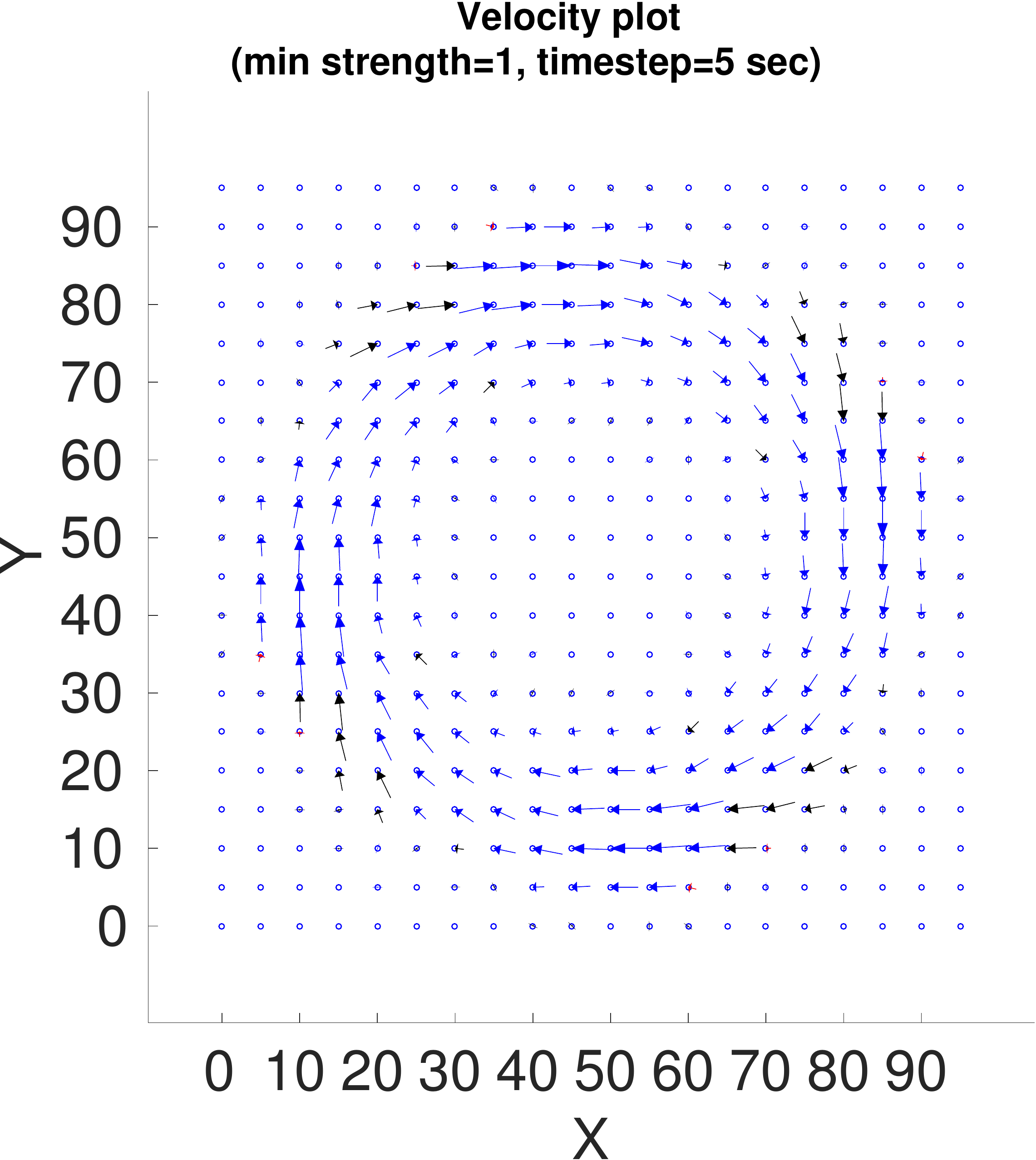}
    \caption{Scenario 2: Ring flow}
\end{subfigure}
\hspace*{3mm}
\begin{subfigure}[t]{.3\textwidth}
	\includegraphics[scale = 0.25,trim=0 0 0 110bp,clip]{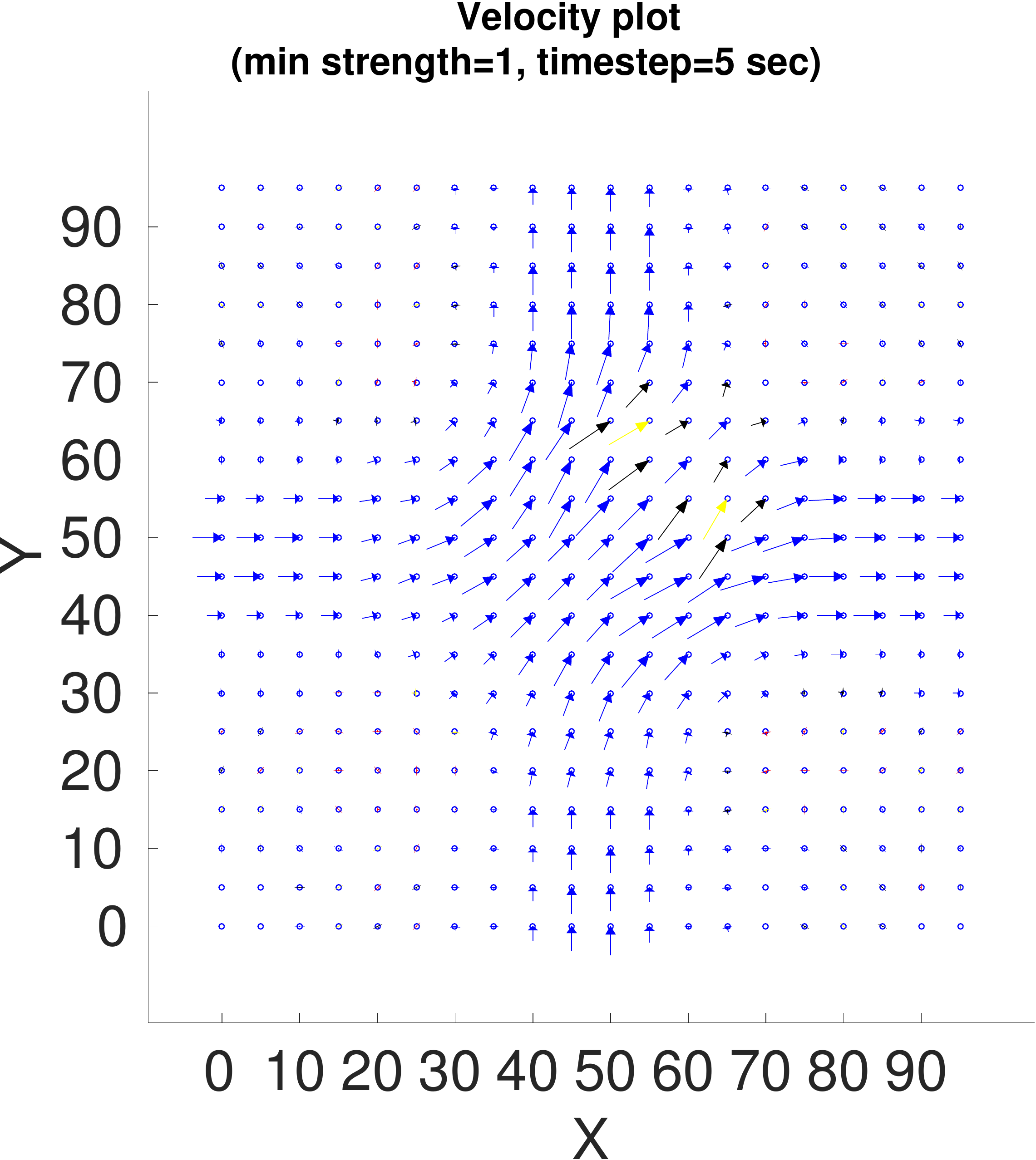}
    \caption{Scenario 3: Cross Current}
\end{subfigure}

\caption{Velocity estimates from PC stable for Scenarios 1, 2 and 3 are decent, but some directions are distorted.}
\vspace{-10mm}
\label{pc_velocity_fig}
\end{figure}

\begin{center}
\begin{table}[t]\label{table-clime}
\centering
\captionsetup{justification=centering}
\caption{Error measures of velocity estimates for synthetic data\\ (CLIME and ACLIME denote CLIME-ADMM and ACLIME-ADMM respectively)} \label{table1}
\begin{tabular}{|l|l||r|r|r|} \hline
Scenario & Method & PPDL15 &  \makecell{RMSE-\\Angle} & \makecell{RMSE-\\Length} \\ \hline\hline
Circular flow & PC stable   & 76 &  27.0671 & 0.8206\\ \hline
  & \makecell{CLIME} & 84 & 21.5059 & 0.9284 \\ \hline
  & \makecell{ACLIME} & 84 &  25.6995 & 0.7994\\ \hline\hline
Ring flow & PC stable       & 90 &  11.2116 & 0.6241\\ \hline
  & CLIME    & 89 & 7.73 &  0.6643 \\ \hline
   & ACLIME    & 83 &  7.1998& 0.6124\\ \hline\hline
Cross Current & PC stable   & 65.5 &  35.7746& 0.7165\\ \hline
   & CLIME & 98.5 & 6.3717 & 0.8277   \\ \hline
   & ACLIME& 100   & 5.1364& 0.7754\\ \hline\hline
Fast Ring Flow & PC stable   & 49 &   59.8538 & 1.5234 \\ \hline
   &CLIME  & failed &  failed & failed \\ \hline
   & ACLIME & 30  & 50.049  & 1.7871  \\ \hline
\end{tabular}
\end{table}
\end{center}

\begin{figure}[ht]

\begin{subfigure}[t]{.3\textwidth}
    \includegraphics[scale = 0.25,trim=0 0 0 110bp,clip]{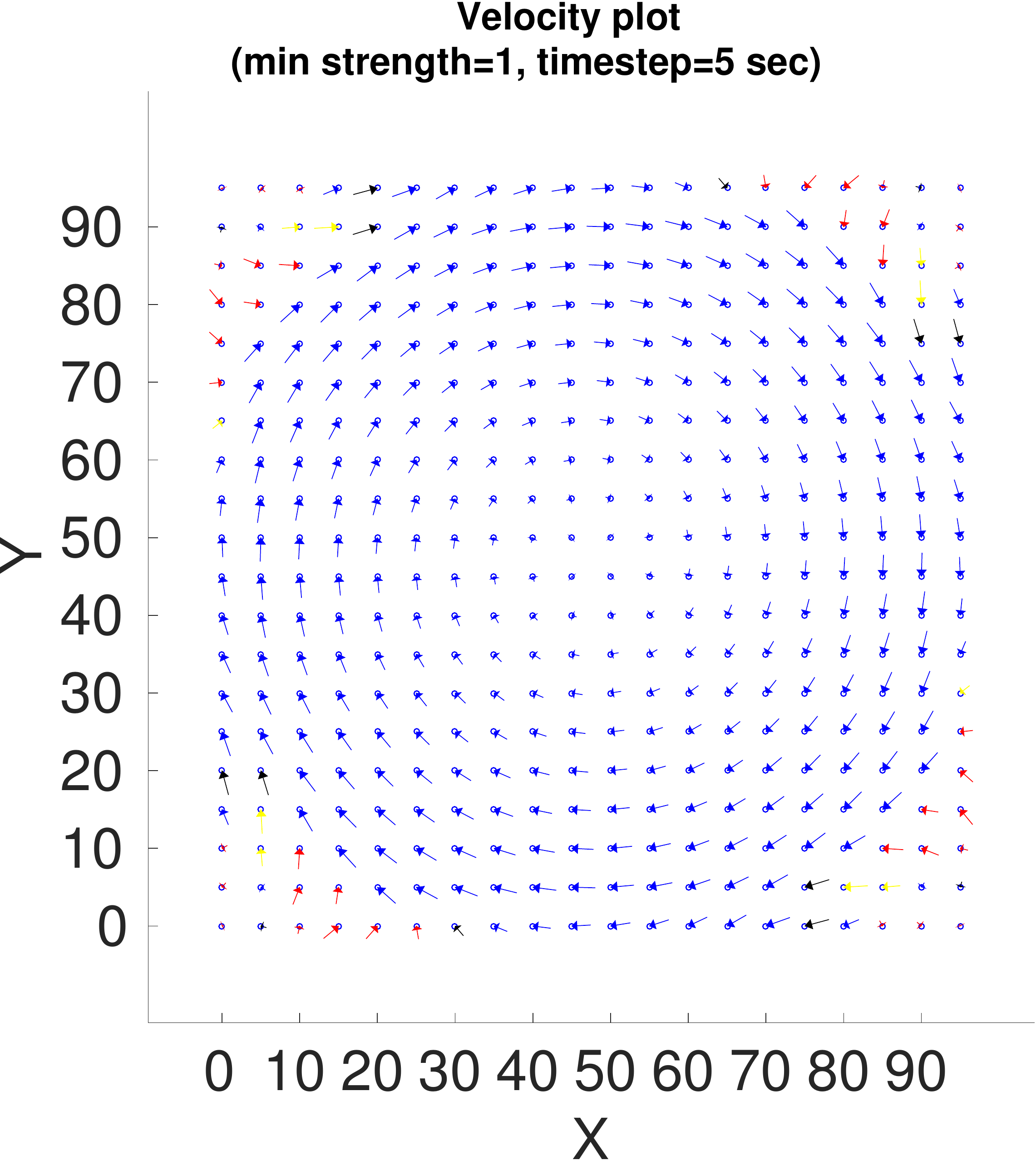}
    \caption{Scenario 1: Circular flow}
\end{subfigure}
\hspace*{3mm}
\begin{subfigure}[t]{.3\textwidth}
    \includegraphics[scale = 0.25,trim=0 0 0 110bp,clip]{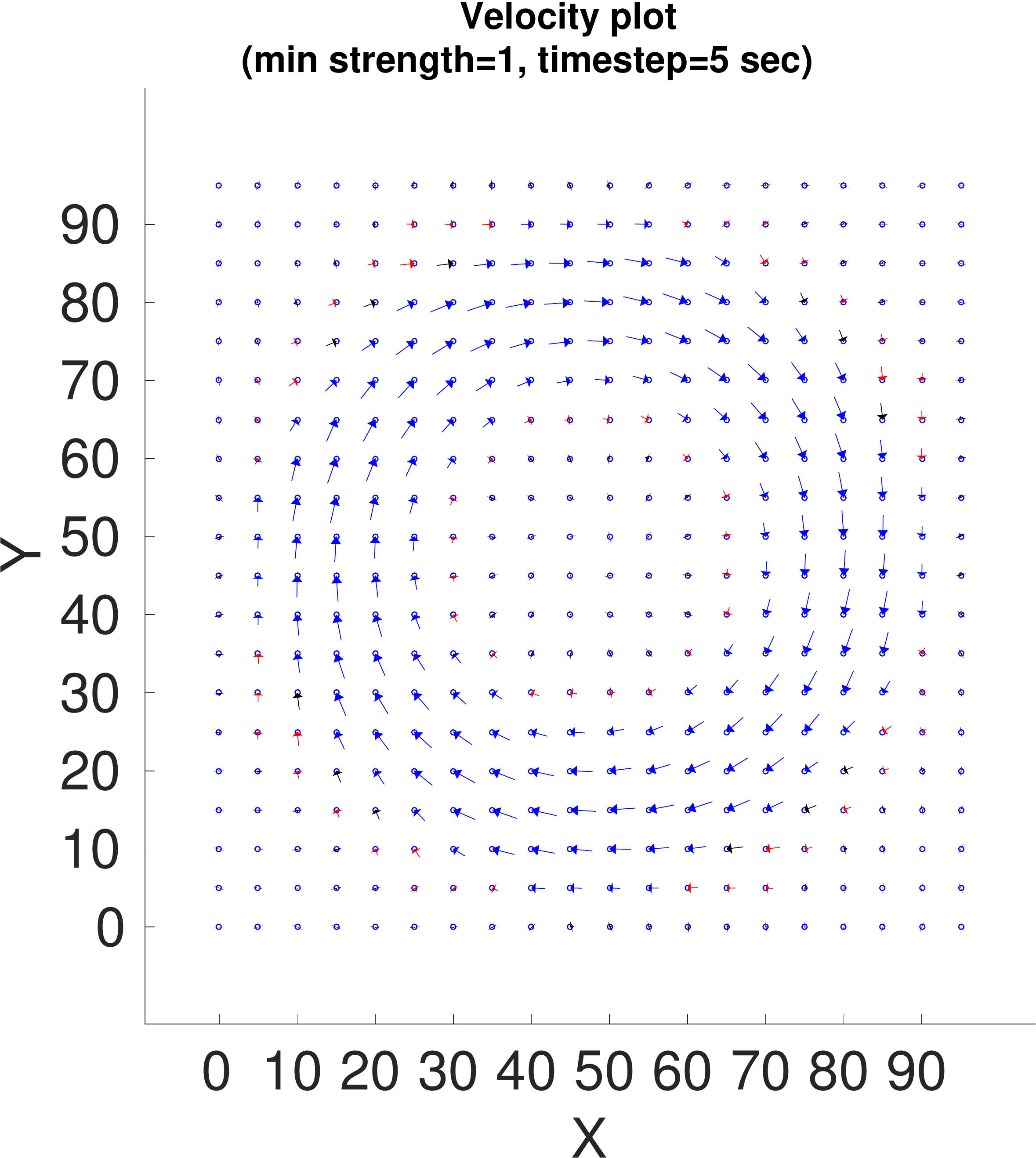}
    \caption{Scenario 2: Ring flow}
\end{subfigure}
\hspace*{3mm}
\begin{subfigure}[t]{.3\textwidth}
	\includegraphics[scale = 0.25,trim=0 0 0 110bp,clip]{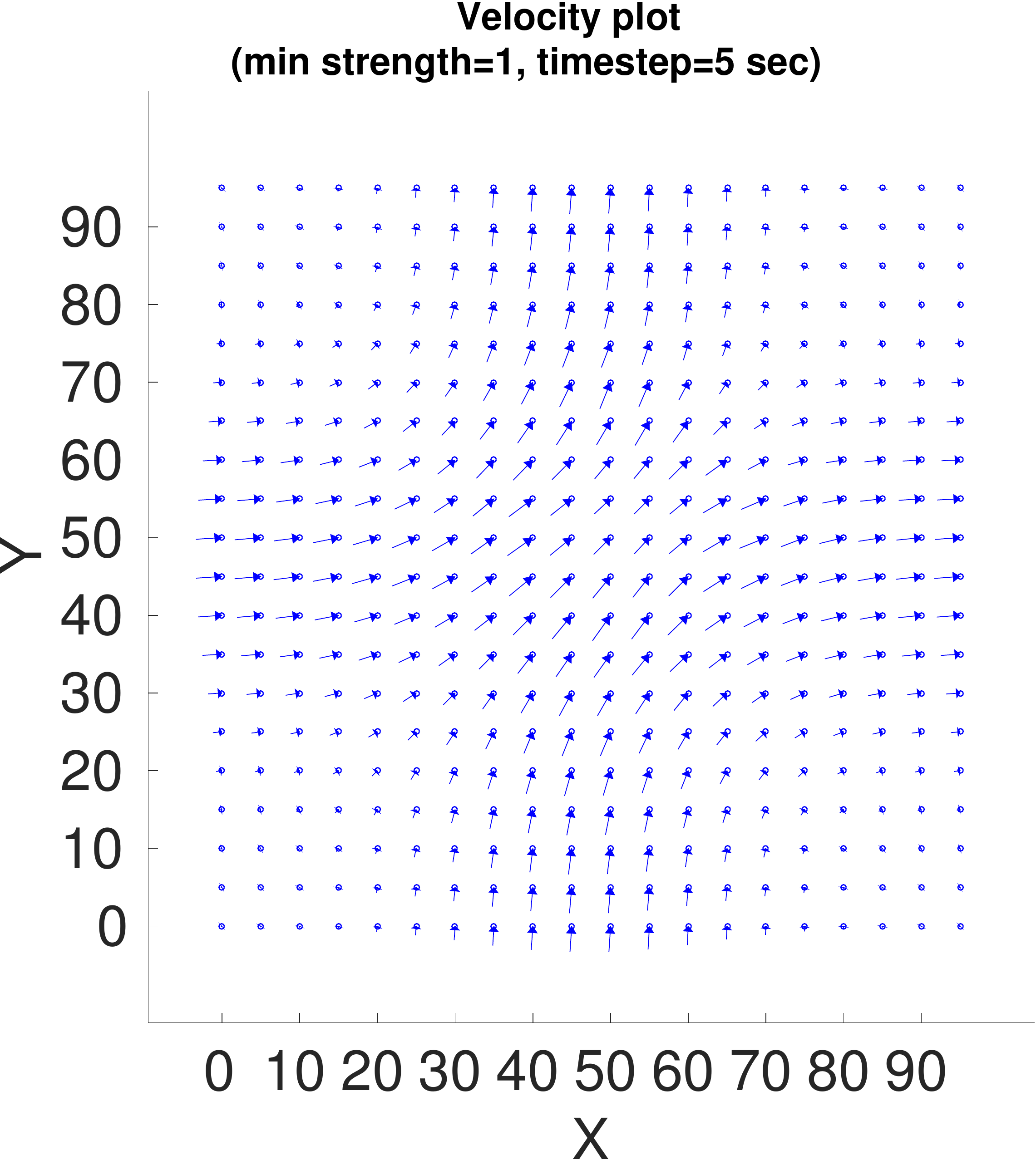}
    \caption{Scenario 3: Cross Current}
\end{subfigure}

\caption{Velocity estimates from ACLIME-ADMM for Scenarios 1, 2 and 3 are more accurate than PC stable.}
\label{clime-velocity-fig}

\end{figure}


\subsection{Results for Synthetic Data.}

\begin{figure}[ht]
\vspace*{-2mm}
\begin{subfigure}[t]{.3\textwidth}
    \includegraphics[scale = 0.25,trim=0 0 0 110bp,clip]{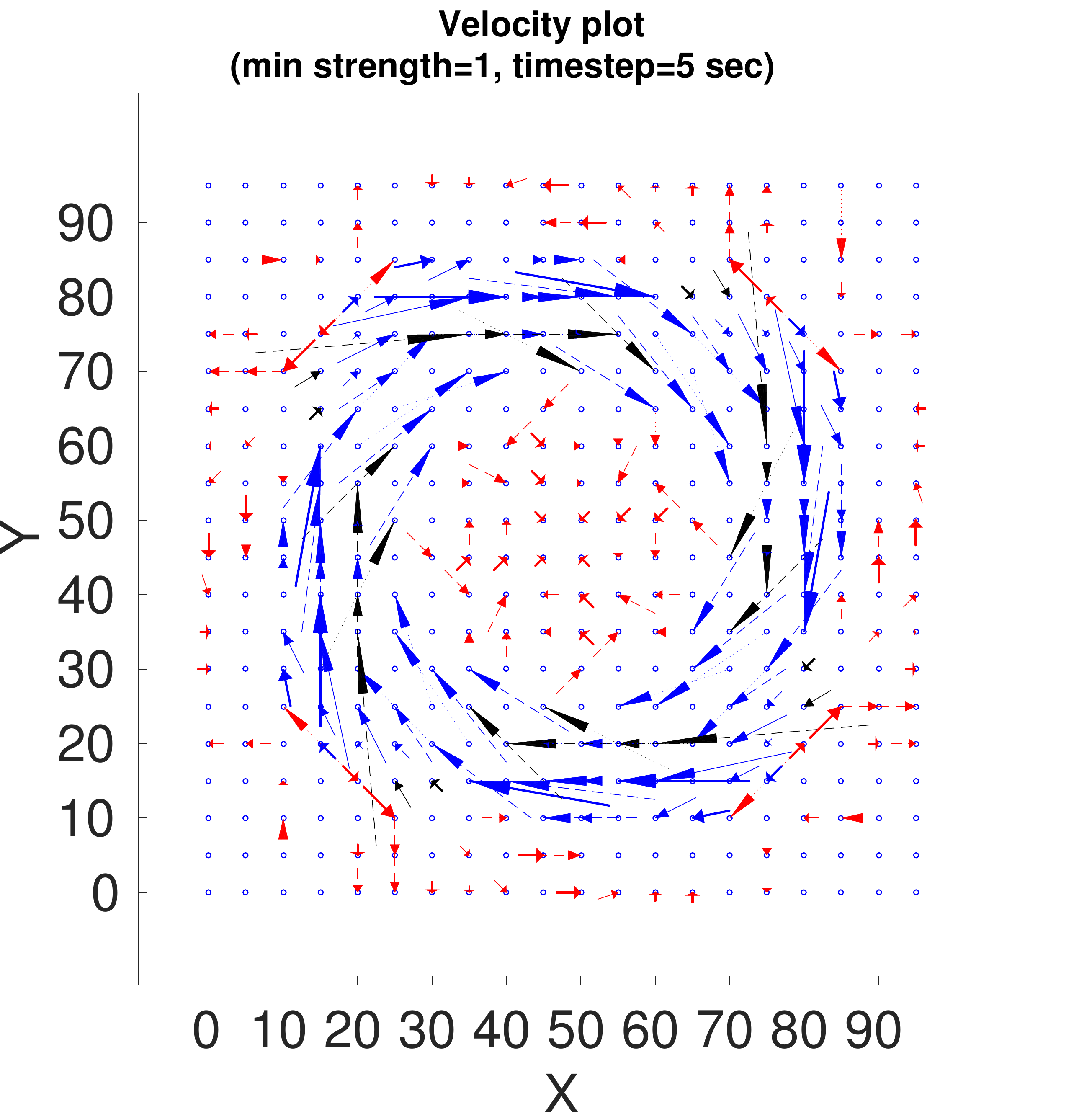}
    \caption{PC stable}
\end{subfigure}
\hspace*{3mm}
\begin{subfigure}[t]{.3\textwidth}
    \includegraphics[scale = 0.19,trim=0 0 0 140bp,clip]{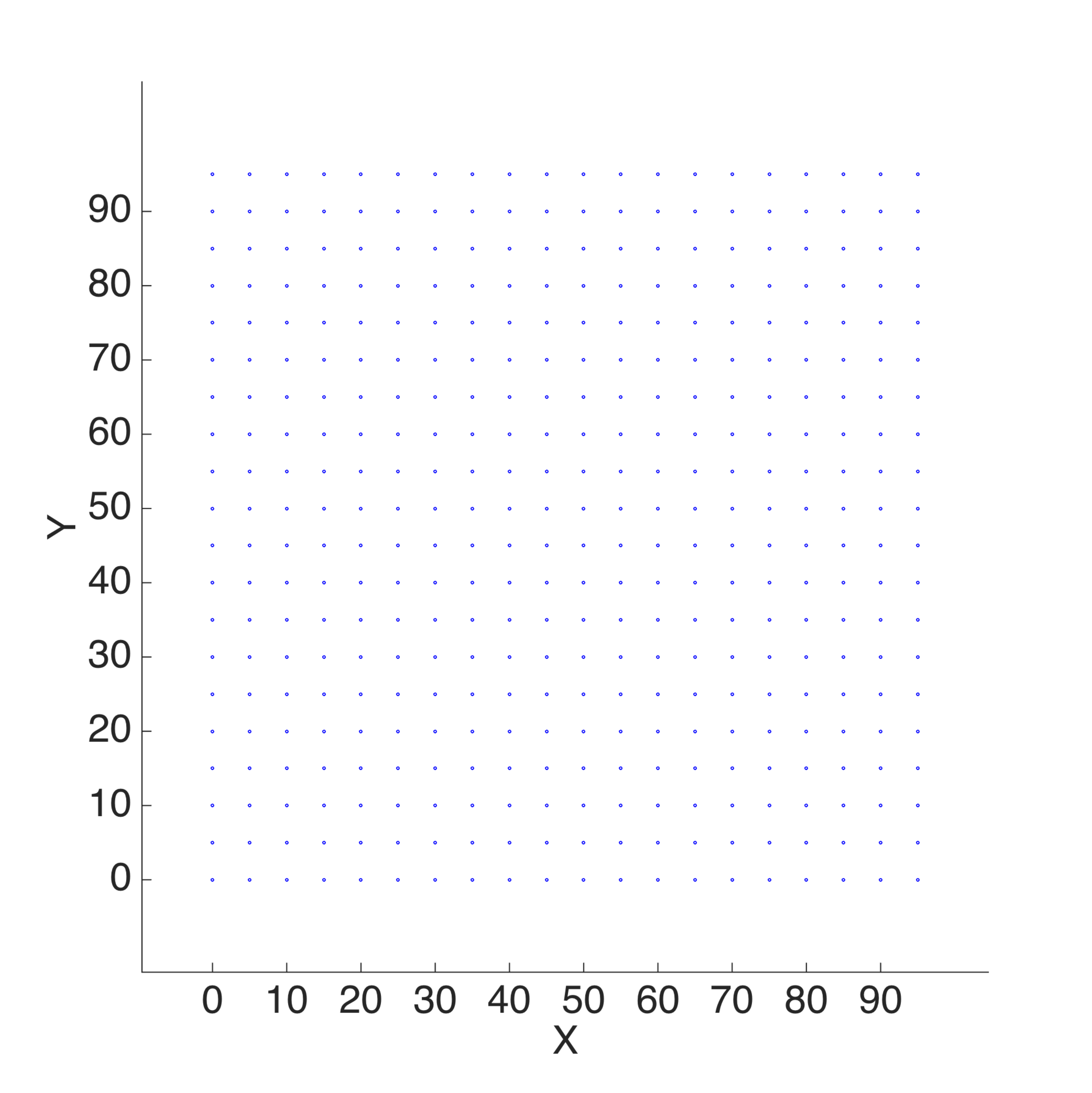}
    \caption{CLIME-ADMM: no velocities found!}
\end{subfigure}
\hspace*{3mm}
\begin{subfigure}[t]{.3\textwidth}
	\includegraphics[scale = 0.25,trim=0 0 0 110bp,clip]{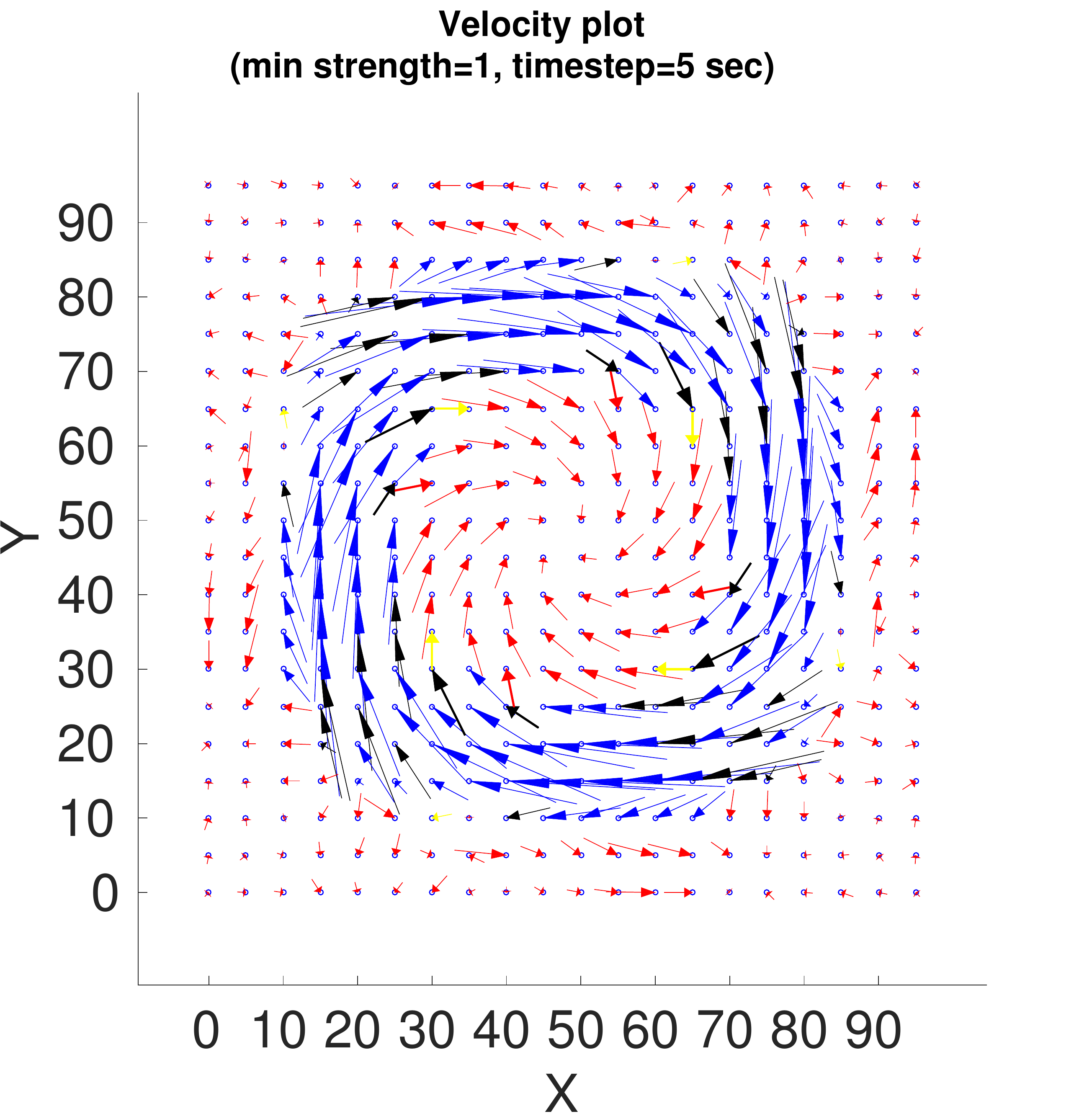}
    \caption{ACLIME-ADMM}
\end{subfigure}
\caption{Velocity estimates from all three algorithms for High-Speed Scenario (Scenario 4, Fast Ring Flow).  PC stable and ACLIME-ADMM detect the high speed signals, but CLIME-ADMM fails miserably, which motivated the development of ACLIME-ADMM in the first place.}
\label{M_10-velocity-fig}
\end{figure}

Fig.\ \ref{pc_velocity_fig} shows the results for the three different scenarios
for the PC stable algorithm and Fig.\ \ref{clime-velocity-fig} for ACLIME-ADMM.
The results from CLIME-ADMM for those three scenarios are very similar to those from
ACLIME-ADMM and are not shown here. Fig.\ \ref{M_10-velocity-fig} shows the results for all three algorithms in one case
where the results actually differ significantly.
Table \ref{table1} shows the error measures for all three algorithms.

Overall, all three algorithms succeed in capturing the main features of the advection fields
for the three main scenarios, but there are some significant differences, discussed below.
\\
{\bf Scenario 1:}
The results for Scenario 1 (Figs.\ \ref{pc_velocity_fig}(a) and \ref{clime-velocity-fig}(a))
highlight several common trends of the algorithms.
ACLIME-ADMM tends to be more sensitive, and is thus better in identifying velocities
of small magnitude, thus there are more edges identified near the center of
Fig.\ \ref{clime-velocity-fig}(a).  However, the PC stable algorithms seems to be
able to better deal with the contradicting edge directions near the boundary of Scenario 1,
as can be seen by the many edges identified correctly near the four corners in Fig.\ \ref{clime-velocity-fig}.  Lastly, the PC stable algorithm generally has a harder time to approximate
connections that are not aligning with the grid (not vertical or horizontal).
As a result the PC stable approximation looks a bit more like a square with rounded corners,
while the ACLIME-ADMM approximation detects an almost perfectly round pattern, which matches
the actual advection velocity field.  The error measures confirm the higher accuracy
of ACLIME-ADMM over PC stable, although the difference is not huge.
%
%
%
%
\\
{\bf Scenario 2:}
Both algorithms do a good job of detecting the ring flow
(Figs.\ \ref{pc_velocity_fig}(b) and \ref{clime-velocity-fig}(b)) and 
identifying zero velocities where appropriate.
Again, PC stable tends to straighten out nearly diagonal edges,
i.e.\ it tends to make them more vertical or horizontal than they should be. Both algorithms do a good job of identifying zero velocities
in the large areas where the advection fields are indeed zero.
According to the error measures PC stable is actually more accurate than ACLIME-ADMM in one measure
for this scenario, PPDL15, but in all other measures, ACLIME performs better
for this scenario. Overall, there is not a huge difference between the two algorithms
for this scenario.
\\
{\bf Scenario 3:}
ACLIME-ADMM truly shines for the cross current scenarios, with the number of arrow directions
identified within a 15 degree error margin at $100\%$
(Fig.\ \ref{clime-velocity-fig}(c)).
In contrast, PC stable struggles more, because of the large number of
diagonal edges in the center of the cross current (Fig.\ \ref{pc_velocity_fig}(c)),
some of which it captures correctly, but others not.
The higher sensitivity of ACLIME-ADMM also helps it better identify velocities throughout
that have smaller magnitude, resulting in a very nice representation of the
original advection field.
Overall, both algorithms perform similarly for length, but ACLIME-ADMM delivers much better results for direction.
\\
{\bf Scenario 4:}
Lastly, we test the algorithms on a modification of Scenario 2. Namely,
we take the simulation data from the ring flow, but only use every 10$th$ sample of the data.
This results in a flow in the same direction, but with signals propagating
at ten times the speed of Scenario 2, which makes them much harder to detect.
Results for this high-speed scenario are shown in Fig.\ \ref{M_10-velocity-fig}.
PC stable shows good results.
However, CLIME-ADMM fails miserably, in fact it does not find a {\it single} connection.
This failure was a primary reason for developing ACLIME-ADMM, namely to provide a
scalable algorithm that can handle high-speed connections.
Indeed, ACLIME-ADMM performs well for this scenario, similarly to PC stable, as seen in
Fig.\ \ref{M_10-velocity-fig}(c).



\remove{

\subsubsection{Scenario 3: Cross Current.}
We present velocity results for Cross Current (scenario 3) for both {\it CLIME-ADMM} and {\it PC stable} (Figure~\ref{velocity-s3}). Both methods capture the cross current pattern, but as before there are a few differences.
{\it PC stable} produces a strange \ab{rephrase?} velocity field near the center where two currents cross each other, whereas the velocity field from {\it CLIME-ADMM} is smooth near the center. Further, the width of the velocity fields captured by {\it PC stable} has a narrower width, whereas the width from {\it CLIME-ADMM} seems qualitatively closer to the synthetic scenario.

} 


\subsection{Results for Observed Data}

\begin{figure}[t]
\begin{center}
\begin{subfigure}{.45\textwidth}
  \includegraphics[scale=0.20]{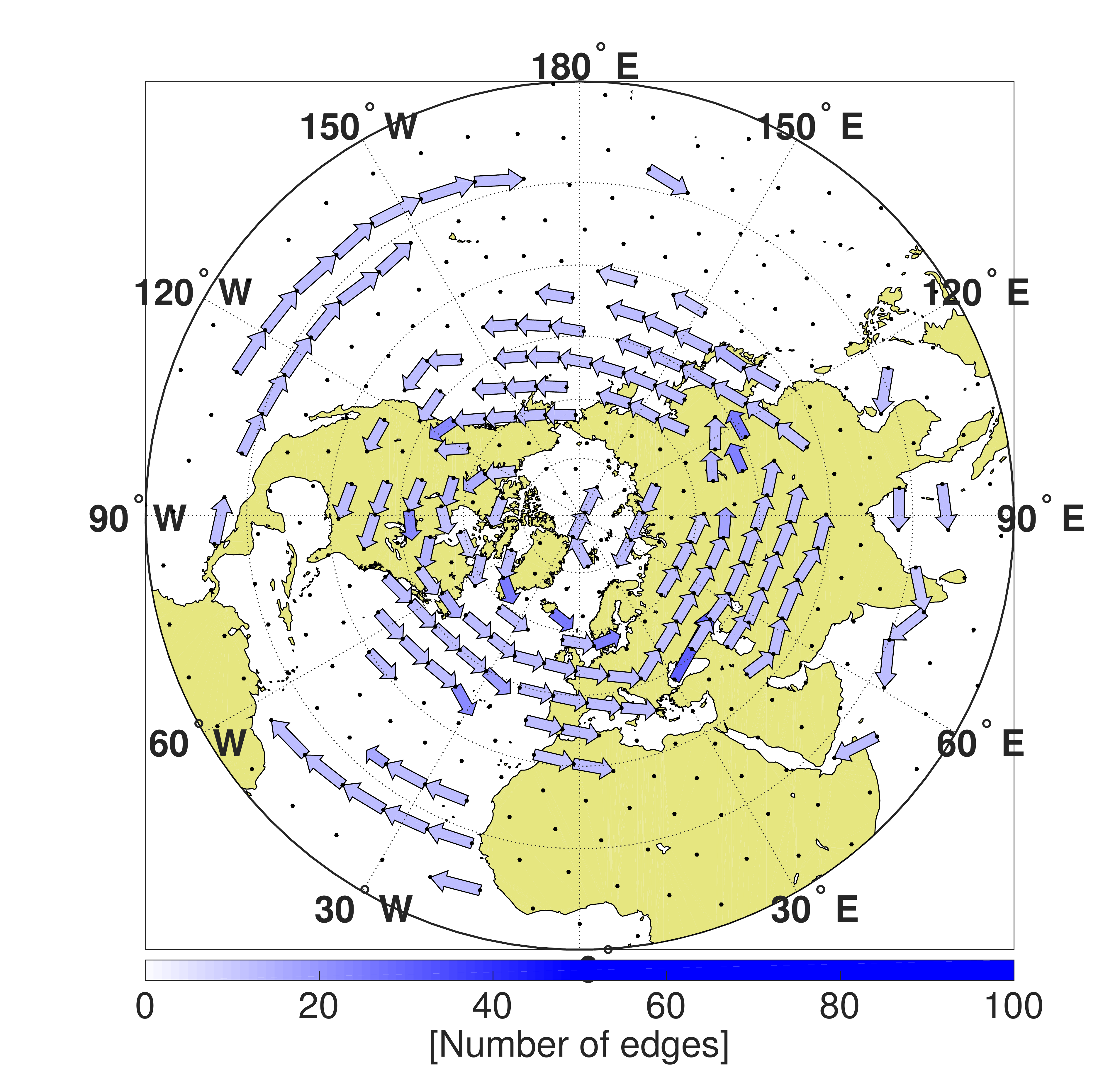}
  \vspace{2mm}
  \caption{PC stable results}
\end{subfigure}
\hspace*{5mm}
\begin{subfigure}{.45\textwidth}
  \includegraphics[scale=0.20]{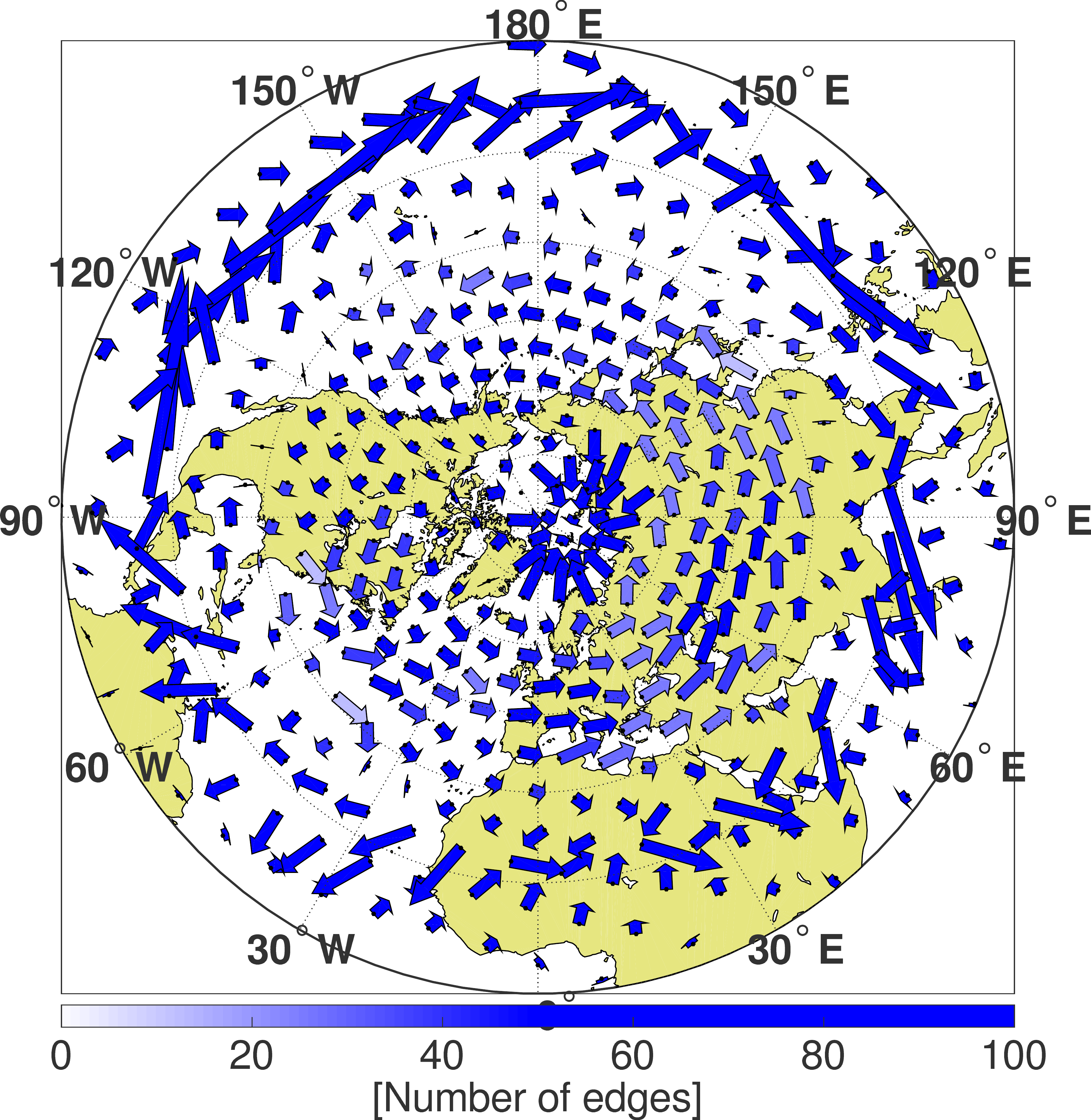}
   \vspace{2mm}
  \caption{ACLIME results}
\end{subfigure}

\caption{Velocity estimates in Northern Hemisphere.}
\label{fig:real-nh}
\vspace*{-0.5cm}
\end{center}
\end{figure}
\begin{figure}[t]
\begin{center}
\begin{subfigure}[t]{.45\textwidth}
\centering
  \includegraphics[scale=0.20]{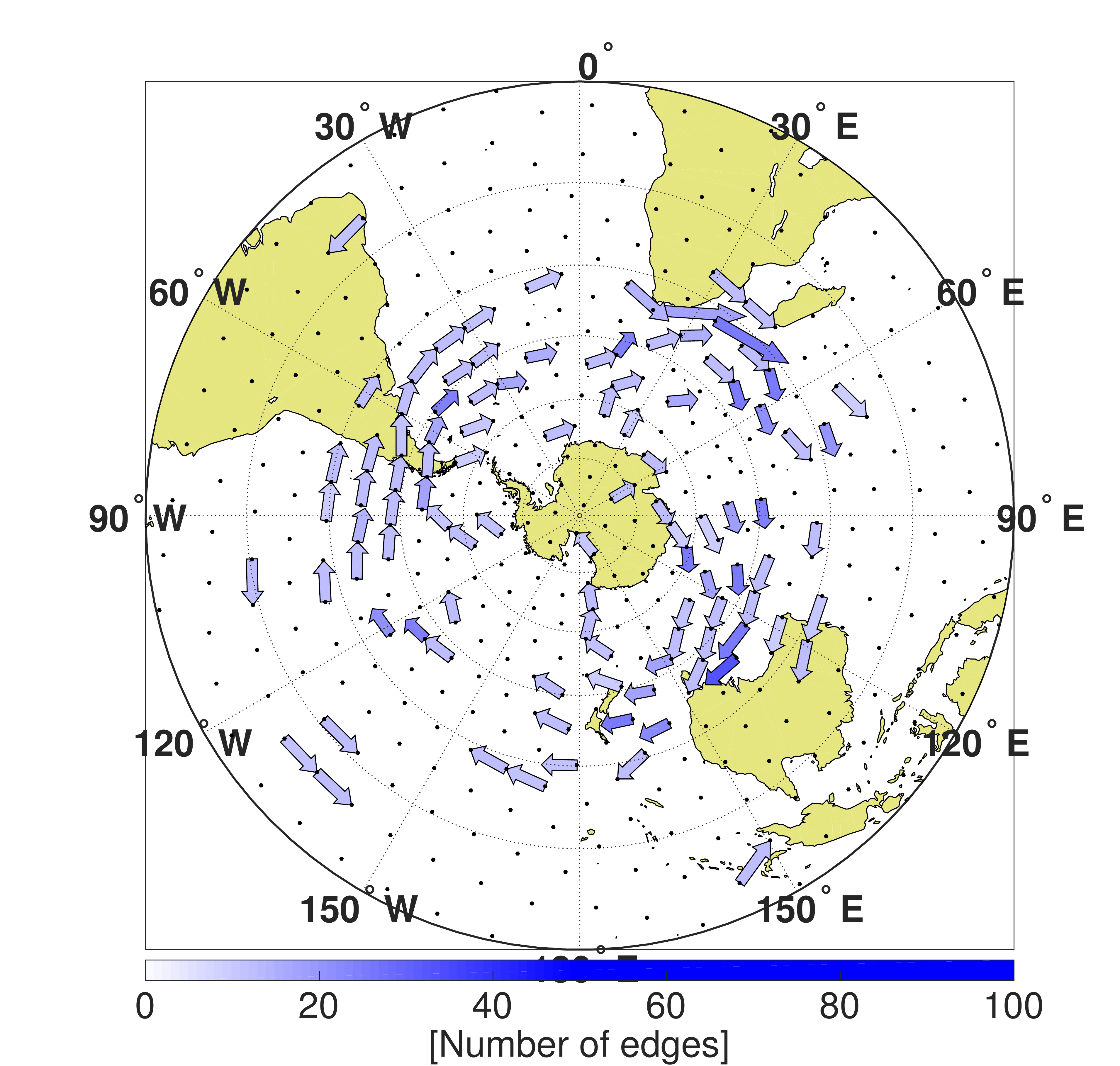}
   \vspace{1mm}
  \caption{PC stable results}
\end{subfigure}%
\hspace*{5mm}
\begin{subfigure}[t]{.45\textwidth}
  \includegraphics[scale=0.20]{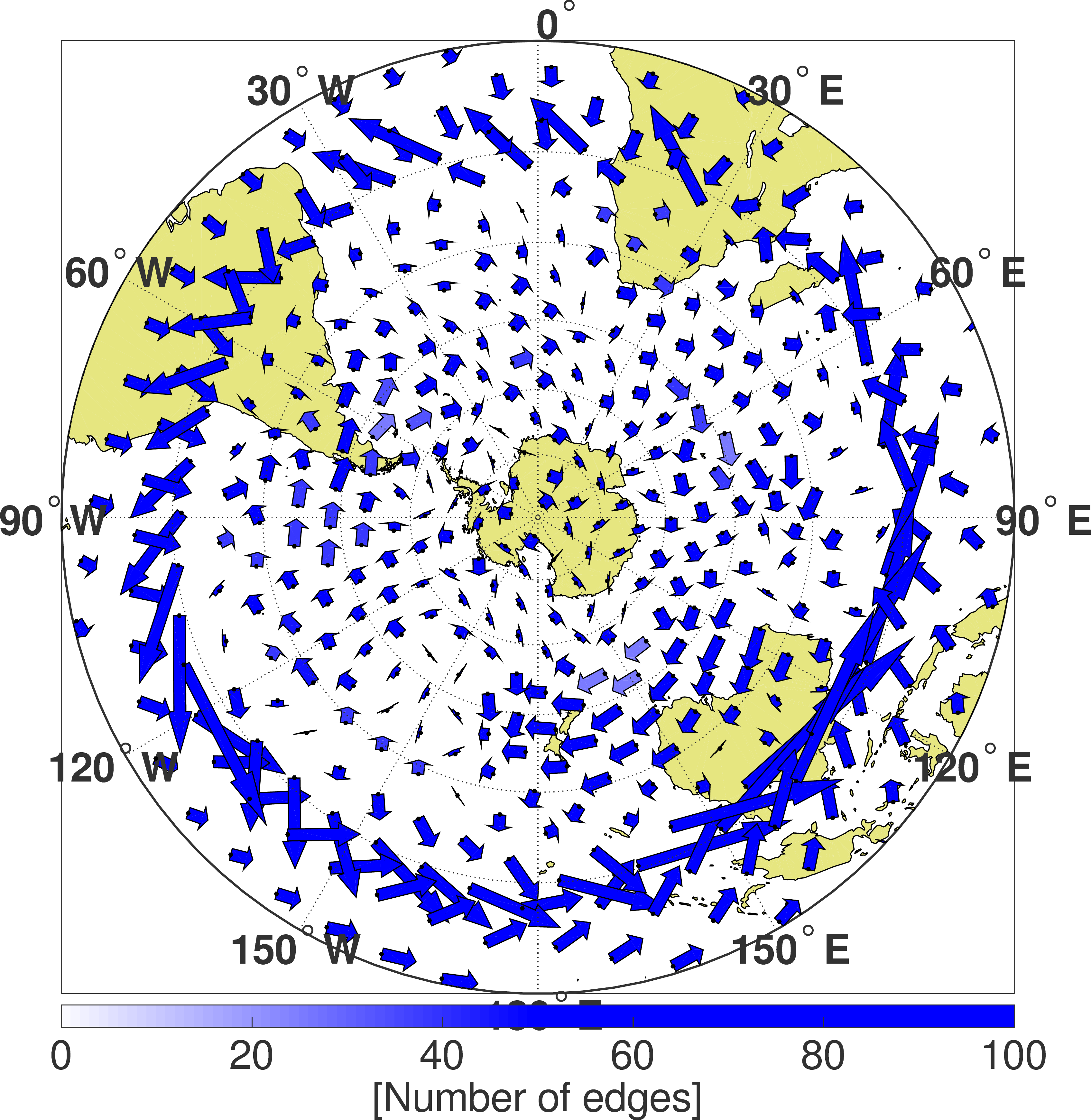}
  \vspace{5mm}
  \caption{ACLIME results}
\end{subfigure}

\caption{Velocity estimates in Southern Hemisphere.}
\label{fig:real-sh}
\vspace*{-0.5cm}
\end{center}
\end{figure}
\begin{figure}
\begin{center}
\begin{subfigure}{.45\textwidth}
  \includegraphics[width=0.9\textwidth]{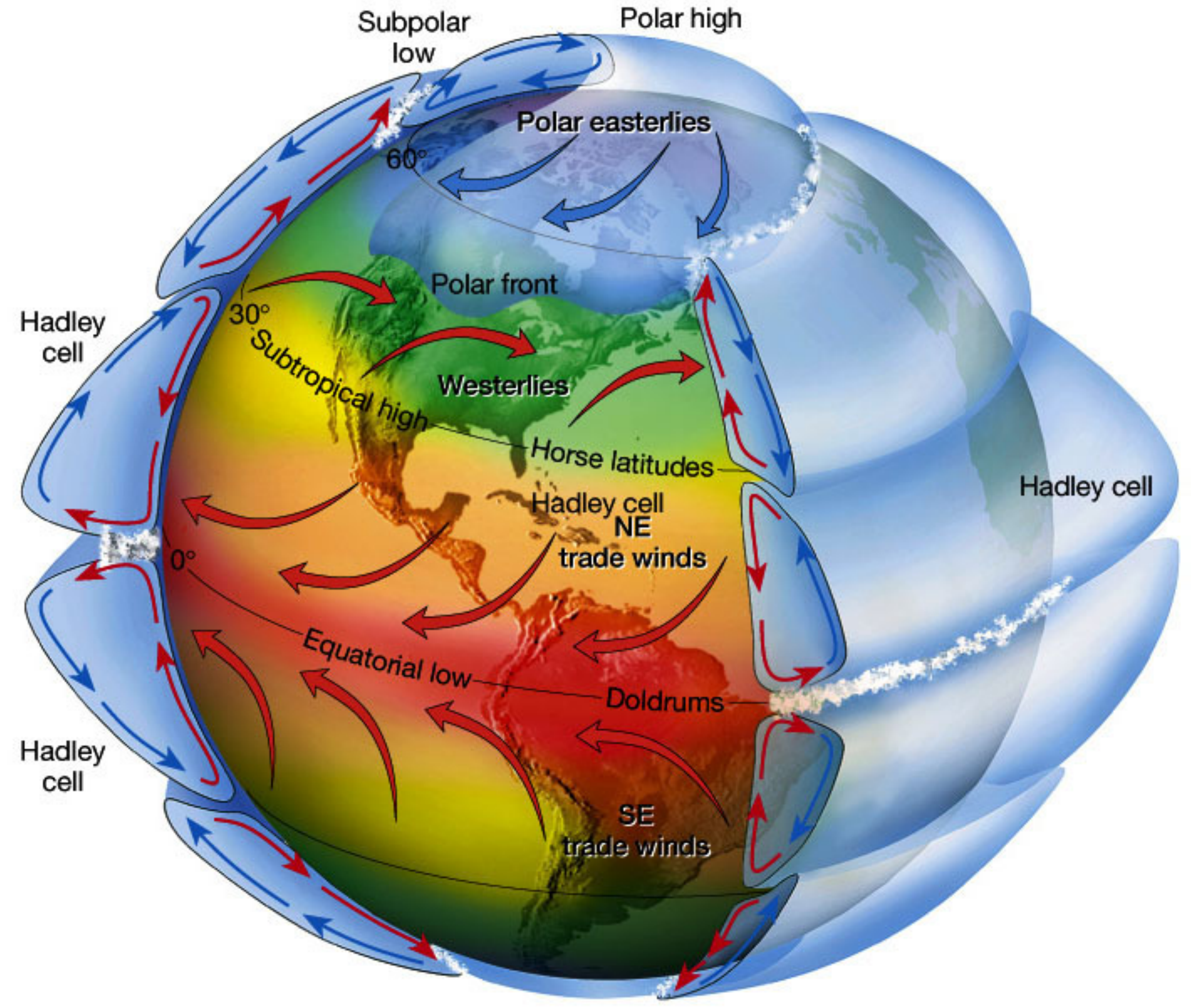}
    \vspace{2mm}
  \caption{Global circulation patterns}
  \label{fig:trade}
  \vspace{3mm}
\end{subfigure}
\begin{subfigure}{.50\textwidth}
 \vspace{2mm}
  \includegraphics[width=0.95\textwidth,trim=0 0 0 75bp,clip]{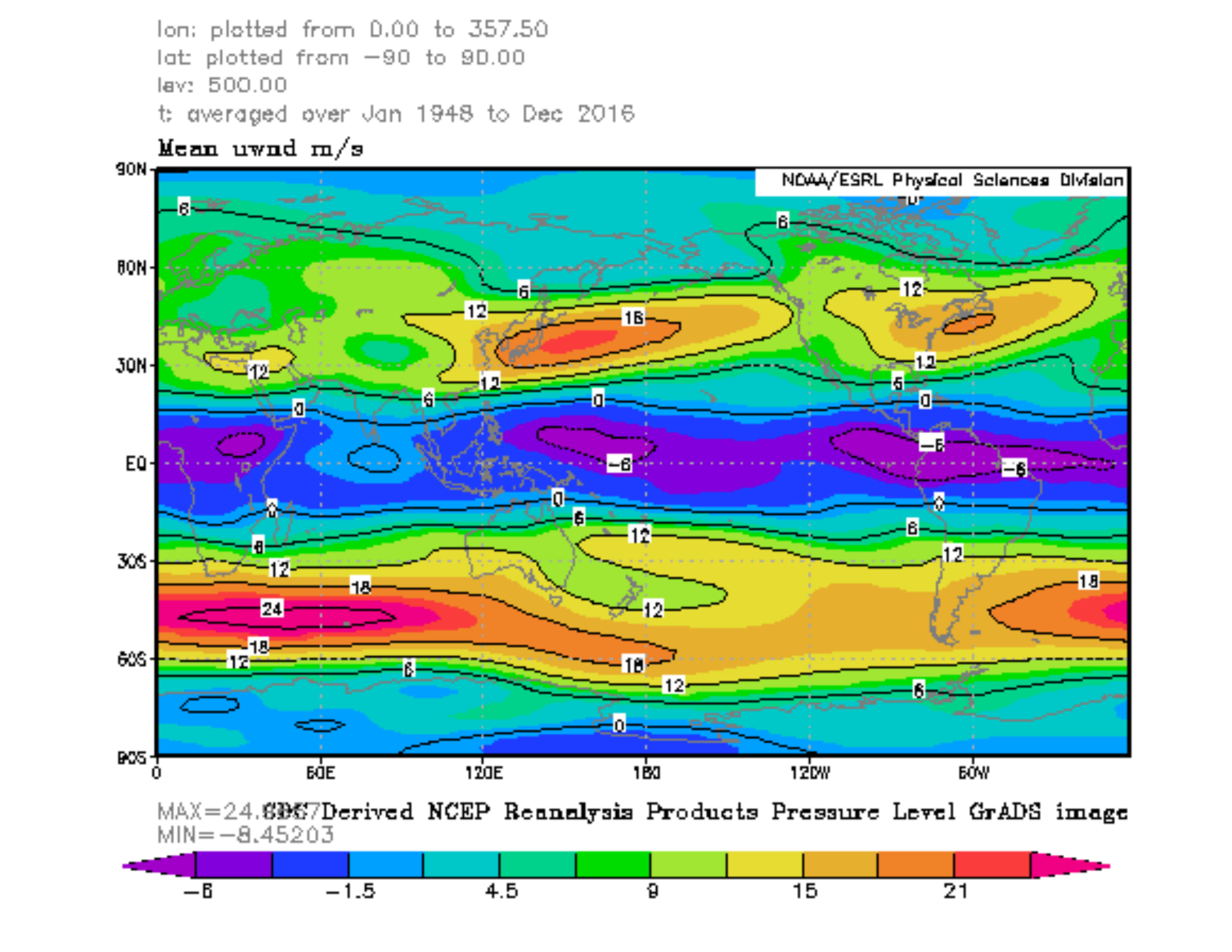}
   \vspace{2mm}
  \caption{Wind at 500mb height}
  \label{fig:wind_at_500mb}
\end{subfigure}
\caption{Atmospheric wind circulation: (a) global circulation patterns and surface winds; and (b) wind at 500 mb height (yearly average).}
\label{fig:wind}
\vspace*{-5mm}
\end{center}
\end{figure}

We compare results from PC stable and ACLIME-ADMM for the dataset of observed
daily geopotential height data (see Section~\ref{ssec:real}).
We show the velocities obtained from both methods in the Northern  (Fig.\ \ref{fig:real-nh}) and Southern hemisphere (Fig.\ \ref{fig:real-sh}).
As a reference, we present the well known wind flow patterns in the Northern and Southern hemispheres in Fig.\ \ref{fig:trade}, as well as wind patterns at 500mb height
in Fig.\ \ref{fig:wind_at_500mb}, which
is the height of the observed data.

The estimates are obtained in a similar way as for the synthetic data, just that
in this case only outgoing edges are considered at each node.
Furthermore, in these plots color is used to indicate connectivity of the grid points.
Namely, for each grid point we count the number of
directed edges incident at that point, i.e.\ the number of edges contributing
to its velocity estimate.
This number indicates strength of connectivity (and thus information flow) at that
point.\footnote{A
consistent connection usually has a strength of at least, say, roughly 10, since it
occurs many times in the adjacency matrix.  For example, a consistent edge
from $P_1$ to $P_2$ with a delay of $4$ time steps occurs in the adjacency matrix as
edge from time stamp 1 to 5, 2 to 6, 3 to 7, etc.
Therefore we only show edges that have a strength of at least 10.
Furthermore, while the color scale reaches its maximum at 100, actual values can be
much higher.}

We make the following observations.
Firstly, ACLIME-ADMM shows even higher sensitivity for the observed data than for the synthetic data, 
resulting in a much larger number of arrows and higher connectivity than PC stable.
Secondly, the results from both algorithms show information transfer
mostly consistent with well known wind directions.
Namely, the spatial distribution of winds at 500mb is such that easterlies
(winds blowing from east to west) dominate the tropical bands (15S-15N),
while westerlies (winds blowing from west to east)
dominate mid latitudes (30N-60N), and another band of weak easterlies are typically seen in the polar region (Figure~\ref{fig:wind}(a)).
PC stable captures the two major bands of easterlies and westerlies, and so does ACLIME-ADMM.  However, the results from ACLIME-ADMM additionally detect very strong information flow near the equator, which cannot be readily explained by the weak easterlies seen at 500mb. 
We are currently exploring alternative explanations, such as these edges maybe being tied to weather features of similar lifecycles occurring simultaneously at different locations,
such as the seasonal march of Intertropical Convergence zone thunderstorms.

\section{Computational Considerations}

Our implementation of PC stable can currently handle up to
about $p=100,000$ nodes, which is the fastest implementation we are aware of.
The biggest limitation to extending the algorithm to more nodes
is not computational time, but working memory.
Running the code for $p=100,000$ nodes requires a computer with
over 100GB of working memory, since
PC stable needs to keep the entire adjacency and covariance matrices in memory
(each is a $p \times p$ matrix),
because it is not known ahead of time which elements will be needed next to
perform conditional independence tests.
Further, although we considered gridded data on a plane (or sphere) for this work, the atmosphere is 3-dimensional, rather than a 2-dimensional, so considering the altitude can drastically increase the number of nodes in the graph, making it more difficult for PC stable.
In contrast ACLIME-ADMM can perform its computations holding only
pre-defined subsets of the adjacency and covariance matrices in memory,
thus has much more promise to scale up to very large numbers of variables.
This fact motivated us to try CLIME-type methods as an alternative in the first place.

%

\section{Conclusions}
\label{sec:conc}


The main contribution of this paper is a new algorithm, ACLIME-ADMM, 
which is suitable for high-dimensional structure learning and for small sample sizes.
The work was motivated by geoscience applications, primarily 
the use of structure learning to
identify interactions between different locations around the globe.
PC stable was previously used for this application and is used here for comparison.
PC stable gives decent, stable results, but is currently limited in the number of variables
it can handle (about 100,000), which is not sufficient for many
high-dimensional geoscience applications extending over both space and time.
CLIME-ADMM, which promises to be much more scalable (already used for 1,000,000 variables
for other applications), was applied for the first time to this application.
It performed well for most scenarios, but failed miserably for the high speed signals (Scenario 4), where PC stable still gave good results.
This motivated the development of the new algorithm, ACLIME-ADMM,
which builds on CLIME-ADMM, but adjusts to
local properties of the data.
ACLIME-ADMM is much more sensitive than PC stable, thus produces denser plots, and is able to identify weaker signals.
For the synthetic data 
ACLIME-ADMM provided the best overall results, including good
results for the high-speed scenario.
For observed data, both algorithms detect the strong easterlies and westerlies bands.
Furthermore,
ACLIME-ADMM yielded
new strong edges near the equator that still need to be traced back to a specific physical mechanism. Clearly, more work needs to be done in order to fully understand the differences between the results obtained from CLIME-ADMM and PC stable. However, ACLIME-ADMM was shown to be a very promising candidate for structure learning in many climate science applications.



\section*{\ackname}

J. Golmohammadi, S. He and A. Banerjee acknowledge the support of NSF grants IIS-1563950, IIS-1447566, IIS-1447574, IIS-1422557, CCF-1451986, CNS-1314560, IIS-0953274, IIS-1029711, NASA grant NNX12AQ39A, and the computing support from the University of Minnesota Supercomputing Institute (MSI). The work was also supported by the NSF Climate and Large-Scale Dynamics (CLD) program through a collaborative grant (AGS-1445956 and AGS-1445978) awarded to Y. Deng and I. Ebert-Uphoff. 

{ 

\bibliographystyle{IEEEtran}
\vspace*{-0.1cm}
\bibliography{sigproc}

\begin{thebibliography}{10}
\providecommand{\url}[1]{#1}
\csname url@samestyle\endcsname
\providecommand{\newblock}{\relax}
\providecommand{\bibinfo}[2]{#2}
\providecommand{\BIBentrySTDinterwordspacing}{\spaceskip=0pt\relax}
\providecommand{\BIBentryALTinterwordstretchfactor}{4}
\providecommand{\BIBentryALTinterwordspacing}{\spaceskip=\fontdimen2\font plus
\BIBentryALTinterwordstretchfactor\fontdimen3\font minus
  \fontdimen4\font\relax}
\providecommand{\BIBforeignlanguage}[2]{{%
\expandafter\ifx\csname l@#1\endcsname\relax
\typeout{** WARNING: IEEEtran.bst: No hyphenation pattern has been}%
\typeout{** loaded for the language `#1'. Using the pattern for}%
\typeout{** the default language instead.}%
\else
\language=\csname l@#1\endcsname
\fi
#2}}
\providecommand{\BIBdecl}{\relax}
\BIBdecl

\bibitem{DeEb:GRL2014}
Y.~Deng and I.~Ebert-Uphoff, ``Weakening of atmospheric information flow in a
  warming climate in the community climate system model,'' \emph{Geophysical
  Research Letters}, vol.~41, no.~1, pp. 193--200, 2014.

\bibitem{chu2005data}
T.~Chu, D.~Danks, and C.~Glymour, ``Data driven methods for nonlinear granger
  causality: Climate teleconnection mechanisms,'' Carnegie Mellon University,
  Tech. Rep., 2005.

\bibitem{EbDe:GRL2012}
I.~Ebert-Uphoff and Y.~Deng, ``A new type of climate network based on
  probabilistic graphical models: Results of boreal winter versus summer,''
  \emph{Geophysical Research Letters}, vol.~39, no.~19, 2012.

\bibitem{IPCC:13}
T.~F. Stocker, D.~Qin, G.-K. Plattner, M.~Tignor, S.~K. Allen, J.~Boschung,
  A.~Nauels, Y.~Xia, V.~Bex, and P.~M. Midgley, ``Climate change 2013: The
  physical science basis,'' Tech. Rep., 2013.

\bibitem{pearl2009causality}
J.~Pearl, \emph{Causality: Models, Reasoning, and Inference}.\hskip 1em plus
  0.5em minus 0.4em\relax Cambridge university press, 2009.

\bibitem{spirtes2000causation}
P.~Spirtes, C.~N. Glymour, and R.~Scheines, \emph{Causation, prediction, and
  search}.\hskip 1em plus 0.5em minus 0.4em\relax MIT press, 2000.

\bibitem{spirtes1991algorithm}
P.~Spirtes and C.~Glymour, ``An algorithm for fast recovery of sparse causal
  graphs,'' \emph{Social science computer review}, vol.~9, no.~1, pp. 62--72,
  1991.

\bibitem{kalisch2007estimating}
M.~Kalisch and P.~B{\"u}hlmann, ``Estimating high-dimensional directed acyclic
  graphs with the pc-algorithm,'' \emph{Journal of Machine Learning Research},
  vol.~8, no. Mar, pp. 613--636, 2007.

\bibitem{harris2013pc}
N.~Harris and M.~Drton, ``Pc algorithm for nonparanormal graphical models.''
  \emph{Journal of Machine Learning Research}, vol.~14, no.~1, pp. 3365--3383,
  2013.

\bibitem{colombo2014order}
D.~Colombo and M.~H. Maathuis, ``Order-independent constraint-based causal
  structure learning.'' \emph{Journal of Machine Learning Research}, vol.~15,
  no.~1, pp. 3741--3782, 2014.

\bibitem{ML-GEOchallenges}
A.~Karpatne, H.~A. Babaie, S.~Ravela, V.~Kumar, and I.~Ebert-Uphoff, ``Machine
  learning for the geosciences - opportunities, challenges, and implications
  for the {ML} process,'' in \emph{SIAM SDM 2017, Workshop on Mining Big Data
  in Climate and Environment}, 2017.

\bibitem{friedman2008sparse}
J.~Friedman, T.~Hastie, and R.~Tibshirani, ``Sparse inverse covariance
  estimation with the graphical lasso,'' \emph{Biostatistics}, vol.~9, no.~3,
  pp. 432--441, 2008.

\bibitem{meinshausen2006high}
N.~Meinshausen and P.~B{\"u}hlmann, ``High-dimensional graphs and variable
  selection with the lasso,'' \emph{The Annals of statistics}, pp. 1436--1462,
  2006.

\bibitem{cai2011constrained}
T.~Cai, W.~Liu, and X.~Luo, ``A constrained {L1} minimization approach to
  sparse precision matrix estimation,'' \emph{Journal of the American
  Statistical Association}, vol. 106, no. 494, pp. 594--607, 2011.

\bibitem{cai2016estimating}
T.~T. Cai, W.~Liu, H.~H. Zhou \emph{et~al.}, ``Estimating sparse precision
  matrix: Optimal rates of convergence and adaptive estimation,'' \emph{The
  Annals of Statistics}, vol.~44, no.~2, pp. 455--488, 2016.

\bibitem{wang2013large}
H.~Wang, A.~Banerjee, C.-J. Hsieh, P.~K. Ravikumar, and I.~S. Dhillon, ``Large
  scale distributed sparse precision estimation,'' in \emph{Advances in Neural
  Information Processing Systems}, 2013, pp. 584--592.

\bibitem{drton2016structure}
M.~Drton and M.~H. Maathuis, ``Structure learning in graphical modeling,''
  \emph{Annual Review of Statistics and Its Application}, no.~0, 2016.

\bibitem{meinshausen2010stability}
N.~Meinshausen and P.~B{\"u}hlmann, ``Stability selection,'' \emph{Journal of
  the Royal Statistical Society: Series B (Statistical Methodology)}, vol.~72,
  no.~4, pp. 417--473, 2010.

\bibitem{Pearl:book}
J.~Pearl, \emph{Probabilistic Reasoning in Intelligent Systems: Networks of
  Plausible Inference}, 2nd~ed.\hskip 1em plus 0.5em minus 0.4em\relax Morgan
  Kaufman, 1988.

\bibitem{chen2006effective}
X.-W. Chen, G.~Anantha, and X.~Wang, ``An effective structure learning method
  for constructing gene networks,'' \emph{Bioinformatics}, vol.~22, no.~11, pp.
  1367--1374, 2006.

\bibitem{liu2012high}
H.~Liu, F.~Han, M.~Yuan, J.~Lafferty, L.~Wasserman \emph{et~al.},
  ``High-dimensional semiparametric gaussian copula graphical models,''
  \emph{The Annals of Statistics}, vol.~40, no.~4, pp. 2293--2326, 2012.

\bibitem{xue2012regularized}
L.~Xue, H.~Zou \emph{et~al.}, ``Regularized rank-based estimation of
  high-dimensional nonparanormal graphical models,'' \emph{The Annals of
  Statistics}, vol.~40, no.~5, pp. 2541--2571, 2012.

\bibitem{banerjee2008model}
O.~Banerjee, L.~E. Ghaoui, and A.~d’Aspremont, ``Model selection through
  sparse maximum likelihood estimation for multivariate gaussian or binary
  data,'' \emph{Journal of Machine Learning Research}, vol.~9, no. Mar, 2008.

\bibitem{hsieh2011sparse}
C.-J. Hsieh, I.~S. Dhillon, P.~K. Ravikumar, and M.~A. Sustik, ``Sparse inverse
  covariance matrix estimation using quadratic approximation,'' in
  \emph{Advances in Neural Information Processing Systems}, 2011, pp.
  2330--2338.

\bibitem{lauritzen1996graphical}
S.~L. Lauritzen, \emph{Graphical models}.\hskip 1em plus 0.5em minus
  0.4em\relax Clarendon Press, 1996, vol.~17.

\bibitem{boyd2011distributed}
S.~Boyd, N.~Parikh, E.~Chu, B.~Peleato, and J.~Eckstein, ``Distributed
  optimization and statistical learning via the alternating direction method of
  multipliers,'' \emph{Foundations and Trends{\textregistered} in Machine
  Learning}, vol.~3, no.~1, 2011.

\bibitem{wang2014bregman}
H.~Wang and A.~Banerjee, ``Bregman alternating direction method of
  multipliers,'' in \emph{Advances in Neural Information Processing Systems},
  2014.

\bibitem{ebert2014causal}
I.~Ebert-Uphoff and Y.~Deng, ``Causal discovery from spatio-temporal data with
  applications to climate science,'' in \emph{Machine Learning and Applications
  (ICMLA), 2014 13th International Conference on}.\hskip 1em plus 0.5em minus
  0.4em\relax IEEE, 2014, pp. 606--613.

\bibitem{kalnay1996ncep}
E.~Kalnay, M.~Kanamitsu, R.~Kistler, W.~Collins, D.~Deaven, L.~Gandin,
  M.~Iredell, S.~Saha, G.~White, J.~Woollen \emph{et~al.}, ``The ncep/ncar
  40-year reanalysis project,'' \emph{Bulletin of the American meteorological
  Society}, vol.~77, no.~3, pp. 437--471, 1996.

\bibitem{bendito2007estimation}
E.~Bendito, A.~Carmona, A.~M. Encinas, and J.~M. Gesto, ``Estimation of fekete
  points,'' \emph{Journal of Computational Physics}, vol. 225, no.~2, pp.
  2354--2376, 2007.

\end{thebibliography}

}
%



\end{document}